\documentclass{article}

\usepackage{microtype}
\usepackage{graphicx}
\usepackage{subfigure}
\usepackage{booktabs} % for professional tables
\usepackage{hyperref}
\usepackage{xcolor}

\usepackage[accepted]{icml2024}

\usepackage{pgfplots}
\usepackage{amsmath}
\usepackage{amssymb}
\usepackage{mathtools}
\usepackage{amsthm}
\usepackage{makecell}
\usepackage{xcolor}
\usepackage{soul}
\usepackage{colortbl}
\usepackage{natbib}
\usepackage{enumerate}
\usepackage{enumitem}
\usepackage{comment}
\usepackage[capitalize,noabbrev]{cleveref}

\usepackage{multirow}
\usepackage{booktabs}
\usepackage{pifont}
\usepackage{xspace}

\theoremstyle{plain}

\theoremstyle{definition}

\theoremstyle{remark}

\definecolor{hlGreen}{HTML}{D9EAD3}
\definecolor{hlPurple}{HTML}{D9D2E9}
\definecolor{hlBlue}{HTML}{C9DAF8}
\definecolor{myRed}{HTML}{dd1c77}
\definecolor{myBlue}{HTML}{2c7fb8}
\definecolor{myGreen}{HTML}{99d8c9}

\newcommand{\pz}{\hphantom{0}}

\newcommand{\hlPurple}[1]{\sethlcolor{hlPurple}\hl{#1}}
\newcommand{\hlBlue}[1]{\sethlcolor{hlBlue}\hl{#1}}
\newcommand{\custompara}[1]{{\vspace{1mm}\noindent\textbf{#1}\xspace}}
\newcommand{\nlp}[1]{\texttt{\footnotesize #1}}

\newcommand{\ModelName}{{MagicLens}\xspace}

\newcommand{\cmark}{\ding{51}}
\newcommand{\xmark}{\ding{55}}

\definecolor{lightgray}{gray}{0.9}
\definecolor{lightlightgray}{gray}{0.95}
\newif\ifarxiv % \arxivtrue, \arxivfalse, and \ifarxiv

\arxivtrue

\usepackage[textsize=tiny]{todonotes}

\icmltitlerunning{MagicLens: Self-Supervised Image Retrieval with Open-Ended Instructions}

\begin{document}

\twocolumn[
%\icmltitle{MagicLens\includegraphics[width=1em]{figures/magiclens.png}: Image Retrieval with Open-ended Instructions}

\icmltitle{MagicLens\includegraphics[width=1em]{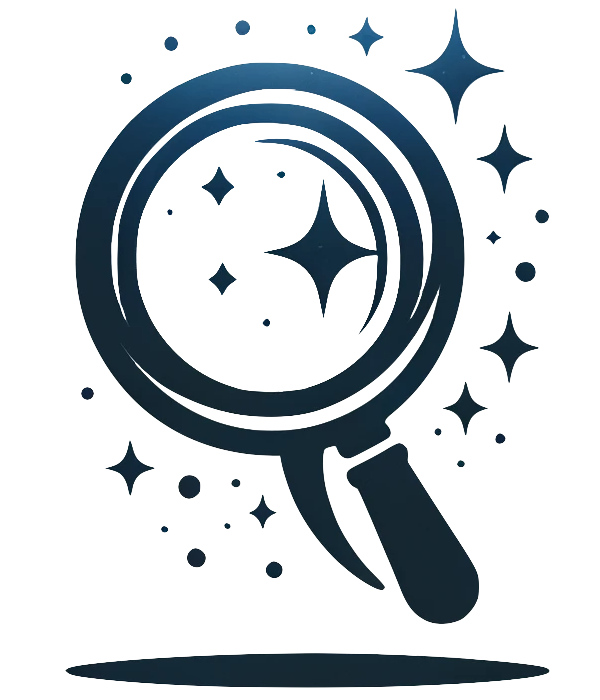}: Self-Supervised Image Retrieval with Open-Ended Instructions}




\icmlsetsymbol{equal}{*}

\begin{icmlauthorlist}
% \icmlauthor{Kai Zhang\textsuperscript{\rm $2\star$}}{}
% \icmlauthor{Yi Luan\textsuperscript{\rm $1$}}{}
% \icmlauthor{Hexiang Hu\textsuperscript{\rm $1$}}{}
% \icmlauthor{Kenton Lee\textsuperscript{\rm $1$}}{}
% \icmlauthor{Siyuan Qiao\textsuperscript{\rm $1$}}{}
% \icmlauthor{Wenhu Chen\textsuperscript{\rm $1$}}{}
% \icmlauthor{Yu Su\textsuperscript{\rm $2$}}{}
% \icmlauthor{Ming-Wei Chang\textsuperscript{\rm $1$}}{} \\
% \textsuperscript{\rm $1$}Google DeepMind~~~
% \textsuperscript{\rm $2$}The Ohio State University
% \\
% \url{https://open-vision-language.github.io/MagicLens/}

\icmlauthor{Kai Zhang}{equal,osu}
\icmlauthor{Yi Luan}{gdm}
\icmlauthor{Hexiang Hu}{gdm}
\icmlauthor{Kenton Lee}{gdm}
\icmlauthor{Siyuan Qiao}{gdm}
\icmlauthor{Wenhu Chen}{gdm}
\icmlauthor{Yu Su}{osu}
\icmlauthor{Ming-Wei Chang}{gdm}
\end{icmlauthorlist}

\icmlaffiliation{osu}{The Ohio State University}
\icmlaffiliation{gdm}{Google DeepMind}

\icmlcorrespondingauthor{Kai Zhang}{zhang.13253@osu.edu}

\icmlkeywords{Machine Learning, ICML}

\vskip 0.3in
]

\printAffiliationsAndNotice{\icmlEqualContribution} % otherwise use the standard text.

\begin{abstract}
% Image retrieval, i.e., finding desired images given a reference image, inherently encompasses rich, multi-faceted search intents that are difficult to capture solely using image-based measures. 
% Recent work leverages text instructions to allow users to more freely express their search intents.
% However, existing work primarily focuses on image pairs that are visually similar and/or can be characterized by a small set of pre-defined relations.
% The core thesis of this paper is that text instructions can enable retrieving images with \textit{richer relations beyond visual similarity}.
% To show this, we introduce \ModelName, a series of self-supervised image retrieval models that support open-ended instructions.
% \ModelName is built on a key novel insight: image pairs that naturally occur on the same web pages contain a wide range of \textit{implicit} relations (e.g., \nlp{inside view of}), and we can bring those implicit relations \textit{explicit} by synthesizing instructions via foundation models.
% Trained on 36.7M (\textit{query image, instruction, target image}) triplets with rich semantic relations mined from the web, \ModelName achieves comparable or better results on eight benchmarks of various image retrieval tasks than prior state-of-the-art (SOTA) methods.
% Remarkably, it outperforms previous SOTA but with a 50$\times$ smaller model size on multiple benchmarks.
% Additional human analyses on a 1.4M-image unseen corpus further demonstrate the diversity of search intents supported by \ModelName.

Image retrieval, i.e., finding desired images given a reference image, inherently encompasses rich, multi-faceted search intents that are difficult to capture solely using image-based measures. 
Recent works leverage text instructions to allow users to more freely express their search intents.
However, they primarily focus on image pairs that are visually similar and/or can be characterized by a small set of pre-defined relations.
The core thesis of this paper is that text instructions can enable retrieving images with \textit{richer relations beyond visual similarity}.
To show this, we introduce \ModelName, a series of self-supervised image retrieval models that support open-ended instructions.
\ModelName is built on a key novel insight: image pairs that naturally occur on the same web pages contain a wide range of \textit{implicit} relations (e.g., \nlp{inside view of}), and we can bring those implicit relations \textit{explicit} by synthesizing instructions via foundation models.
Trained on 36.7M (\textit{query image, instruction, target image}) triplets with rich semantic relations mined from the web, \ModelName achieves results comparable with or better than prior best on eight benchmarks of various image retrieval tasks, while maintaining high parameter efficiency with a significantly smaller model size.
Additional human analyses on a 1.4M-image unseen corpus further demonstrate the diversity of search intents supported by \ModelName.
Code and models are publicly available at the \href{https://open-vision-language.github.io/MagicLens/}{Project Website}.

\end{abstract}

\section{Introduction}
\begin{figure}[t!]
    \centering
    % \vspace{-5mm}
    \includegraphics[width=0.99\linewidth]{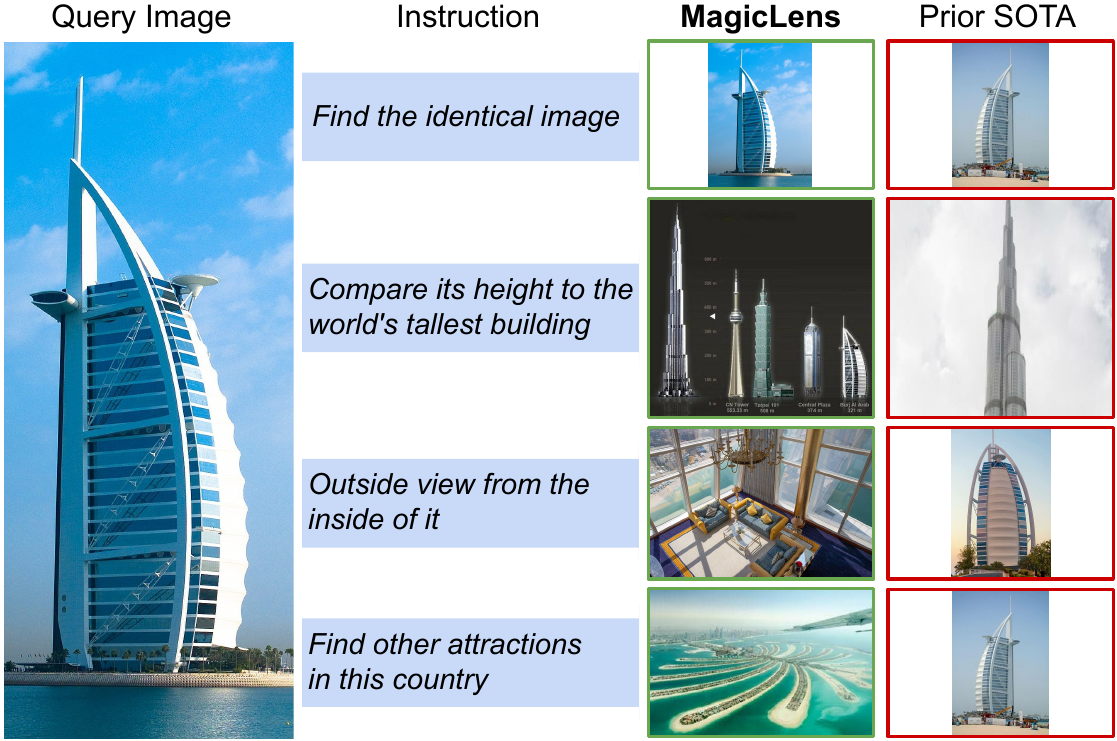}
    \vspace{-1.5mm}
    \caption{
    Top-1 retrieved images using \ModelName and the prior state-of-the-art (SOTA) method~\cite{Gu2023LinCIR} from a retrieval pool with 1.4M images.
    The prior SOTA method, while accepting text instructions, primarily retrieves images based on visual similarity to the query image, ignoring the nuances of the text instructions. In contrast, \ModelName excels at retrieving both visually similar images and those that align with the deeper meaning and context of the text instructions --- even when the images do not resemble the query. For example, if given a query image of the Burj Al Arab and the instruction ``Find other attractions in this country'', it can successfully locate images of the Palm Islands in Dubai.
    }
    \vspace{-4mm}
    \label{fig:teaser-figure}
\end{figure}

Image retrieval is a long-established problem in computer vision~\cite{Datta2008ImageRetrieval, Gordo2016DeepImageRetrieval} with a wide range of real-world applications, such as visual search, object localization, and re-identification.
However, since its inception, this task has suffered from ambiguous definitions due to the complex and rich content encapsulated in images. Similar images may differ in key aspects, and different images can share commonalities. In image search scenarios, users frequently present multiple search intents for a single query image, indicating that mere image relevance is insufficient for precise search results. For instance, when searching with an image of the \texttt{Burj Al Arab hotel in Dubai} (see Figure~\ref{fig:teaser-figure}), a user might seek other attractions in Dubai or an interior view, each relating differently to the query image. Therefore, incorporating text instructions that articulate search intents is essential and indispensable for enhancing retrieval accuracy.
Ideally, models should accurately capture and interpret diverse real-world search intents as conveyed by \textit{open-ended text instructions}.

These open-ended search instructions, span a wide range of topics and concepts, and reflect the diverse ways users interact with visual content, requiring the retrieval system to grasp not only the visual features of an image but also the nuanced semantic relation between the query image and desired results as expressed in the instructions.
Existing models, however, are optimized towards one or a few restricted domains~\cite{Vo2019CIR, Wu2021FashionIQ, Liu2021CIRR, Baldrati2023CIRCO}, where the types of visual similarities are manually defined as \textit{a prior}.
They either adjust the model architecture and training recipe to utilize image-caption data~\cite{Chen2023PLI, Saito2023Pic2Word, Baldrati2023CIRCO, Gu2023LinCIR}, or rely on synthetic data constructed from pre-defined instruction templates~\cite{Brooks2023InstructPix2Pix, Gu2023CompoDiff}.
As a result, neither of these research directions can effectively model open-ended instructions, evidenced by Figure~\ref{fig:teaser-figure}.

\begin{figure}[tb]
    \centering
    % \vspace{-1mm}
    \includegraphics[width=0.96\linewidth]{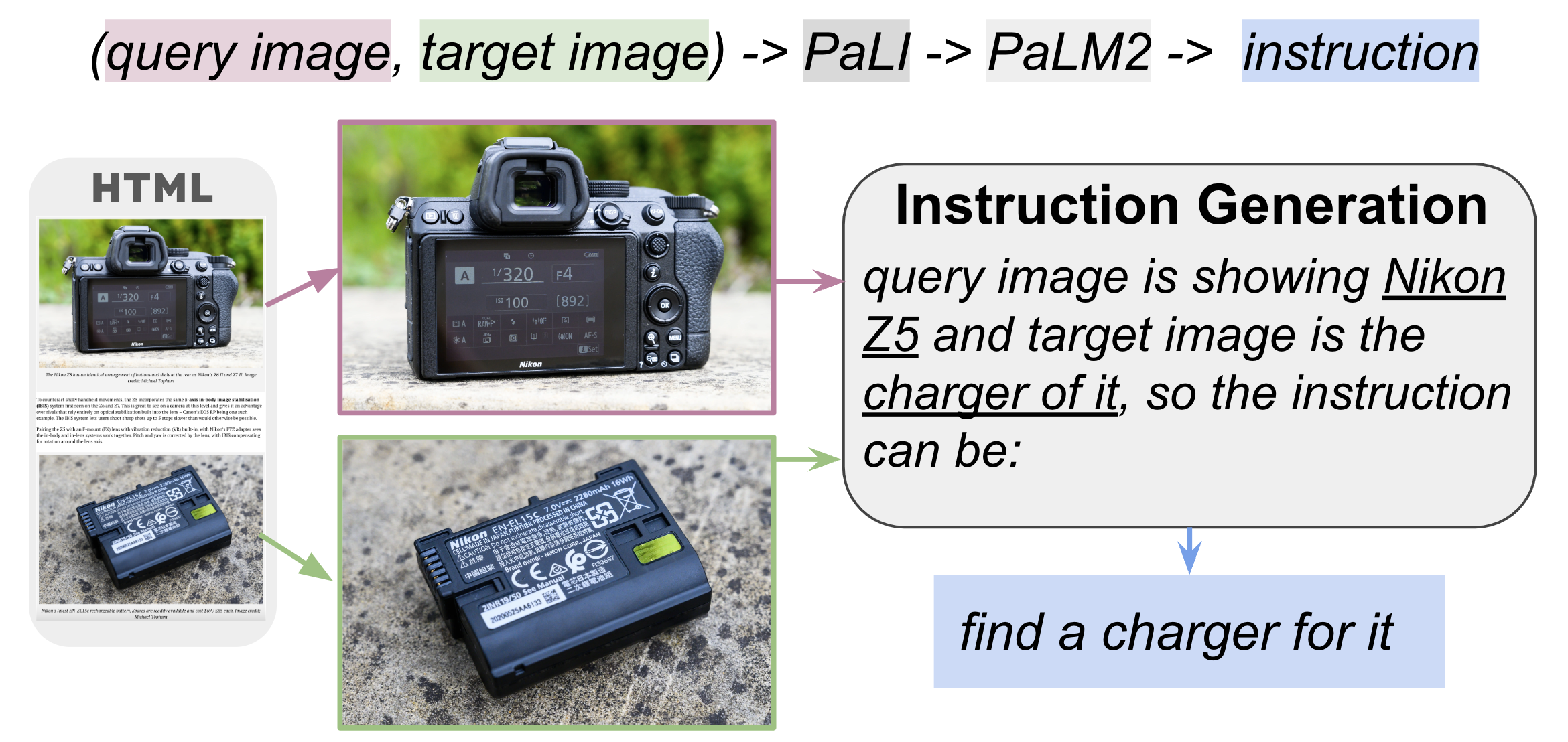}
    % \vspace{-5mm}
    \caption{Data construction overview.
    We collect \textit{\textbf{naturally occurring image pairs}} from the same web pages and use PaLI+PaLM2 to generate instructions connecting the two images.}
    \vspace{-5mm}
    \label{fig:data-construction-overview}
\end{figure}

In this paper, we present \ModelName, a series of self-supervised image retrieval models trained on a wide range of (\nlp{query image}, \nlp{instruction}, \nlp{target image}) triplets that reflect \textit{naturally occurring} semantic relations, mined from web pages and curated with state-of-the-art (SOTA) foundation models. Specifically, we extract image pairs that \textit{naturally occur} on the same web page to form positive pairs that carry abundant but natural semantic relations. We then apply both large multimodal models~(LMMs; \citet{Chen2023PaLI,chen2023palix}) and large language models~(LLMs; \citet{Anil2023PaLM2}) to refine the description of such open-ended semantic relation, into open-ended instruction. Figure~\ref{fig:data-construction-overview} shows an overview of the data construction pipeline. For example, a camera review website\footnote{https://amateurphotographer.com/review/nikon-z5-review/} presenting the image of a \nlp{Nikon Camera} and the image of a \nlp{Nikon Charger} would offer an interesting and non-trivial relation \textit{``charger of a product''}, which would then be curated by the LMM$+$LLM pipeline, and produce a final instruction \nlp{find a charger for it}. This process produces open-ended instructions that depict diverse semantic relations beyond mere visual similarity, resulting in a large-scale training dataset with 36.7M high-quality triplets over a wide distribution.

With the constructed dataset, we train dual-encoder models called \ModelName, which retrieve images given a query consisting of an image with an instruction.
Our models achieve results comparable with or better than prior SOTA methods on eight benchmarks, including various multimodality-to-image and image-to-image retrieval tasks.
In addition, \ModelName can retain or even significantly improve the text-to-image retrieval performance of the underlying single-modality encoders.
With a 50 times smaller model size than prior SOTA methods, \ModelName outperforms them on multiple benchmarks: CIRCO~\cite{Baldrati2023CIRCO}, Domain Transfer ImageNet~\cite{Saito2023Pic2Word}, and GeneCIS~\cite{Vaze2023GeneCIS}.
To further examine our models' capabilities in a more realistic scenario, we construct the largest retrieval pool to date with 1.4 million unseen images and perform retrieval given human-written search queries with diverse instructions.
The human evaluation finds that \ModelName can successfully satisfy complex and beyond visual search intents, whereas the prior SOTA fails to do so.

\begin{figure*}[tbh]
    \centering
    \includegraphics[width=0.96\textwidth]{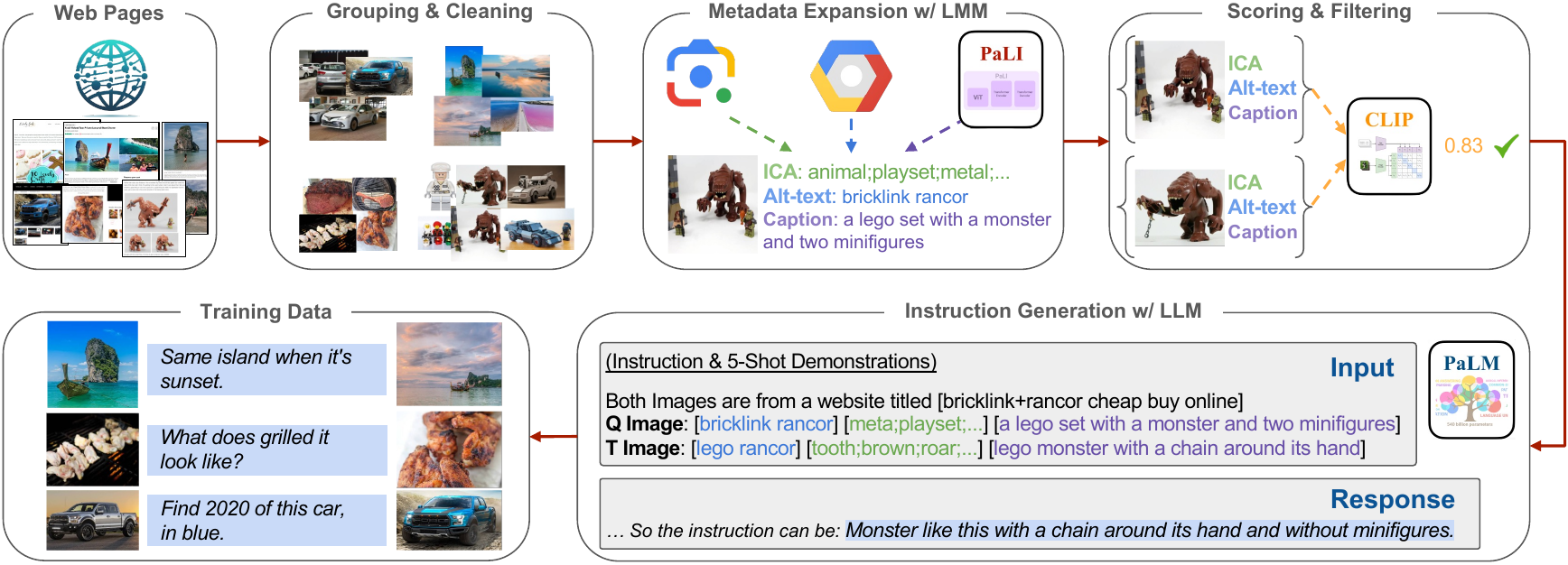}
    \vspace{-2.5mm}
    \caption{Data construction pipeline.
    We mine image pairs from the web via (1) grouping images from the same web page and cleaning them, (2) annotating metadata for each image with LMMs, and (3) scoring and filtering out unqualified image pairs. Eventually, we generate open-ended instructions using LLMs for the remaining image pairs.}
    \vspace{-3mm}
    \label{fig:data-construction}
\end{figure*}

\textbf{Our contributions} are threefold:
\begin{itemize}[leftmargin=*, nosep]
    \item We bring a novel insight for image retrieval: naturally occurring image pairs from the same web pages are strong self-supervised training signals. Based on this, we propose an effective pipeline, backed with LMMs and LLMs, to construct training data consisting of 36.7M triplets.
    \item We introduce \ModelName, a series of light-weight dual-encoders that jointly embed a pair of image and instruction, trained on the constructed dataset.
    Across multiple benchmarks, \ModelName outperforms previous SOTA methods but with a 50$\times$ smaller model size.
    \item We conduct an in-depth human evaluation and analysis on a 1.4M-scale retrieval pool, which is the largest to date. Remarkably high success rates show that \ModelName can well capture and satisfy diverse search intents, especially complex and beyond visual ones.
    % \item We open-source the CLIP-based \ModelName checkpoints and inference code.
    % Their availability can help the development of new image retrieval applications and potentially benefit other vision-language tasks such as visual QA and retrieval-augmented models.
\end{itemize}

\section{Related Work}
\label{sec:related-work}

\custompara{Pre-Training Multimodal Encoders.}
Multimodal encoder pre-training~\cite{faghri2017vse++, chen2021learning, Radford2021CLIP, Yu2022CoCa, Li2021ALBEF, Kim2021ViLT, Wang2023BEiT3, Li2022BLIP, Li2023BLIP2, Cherti2023OpenCLIP} has witnessed great success in recent years.
Pre-trained on web-scale image-caption data~\cite{Zhai2022JFT, Schuhmann2022LAION5B}, these models align the representations of different modalities in a joint space, enabling zero-shot cross-modality retrieval.
However, these works focus on encoding single modalities, without considering the composed representation of multiple modalities.
Some later efforts~\cite{hu2023open,Chen2023InfoSeek,Wei2023UniIR} attempt to combine text and image embeddings via fine-tuning a small number of parameters on top of pre-trained single-modal encoders, without large-scale joint pre-training.
Consequently, such an adaptation strategy shows inferior results on the task of our interest, emphasizing the importance of the \ModelName' self-supervised training.

\custompara{Composed Image Retrieval.}
Composed image retrieval (CIR;~\citet{Vo2019CIR}) shares the same task form with us.
However, all existing benchmarks~\cite{Liu2021CIRR, Baldrati2023CIRCO, Wu2021FashionIQ} collect visually similar images first and then write instructions for image pairs.
This limits the richness of image relations on these benchmarks and the models developed upon/for them.
Recent works on zero-shot CIR~\cite{Saito2023Pic2Word, Baldrati2023CIRCO, Gu2023LinCIR} either design light-weight modality transformation or adjust training and model to use existing image-caption data.
CIReVL~\cite{Karthik2023CIReVL} uses LLM and LMM on-the-fly for CIR, limiting its efficiency.
Please refer to Appendix~\ref{appendix:baselines} for more details of these methods.
In terms of constructing training data, CompoDiff~\cite{Gu2023CompoDiff} synthesizes 18M triplets with LLMs and image generative models, following the same pipeline with~\citet{Brooks2023InstructPix2Pix}.
The key difference between their data and ours lies in the image quality and the image relations.
As shown in Figure~\ref{fig:data-construction-overview}, our data comes from natural image pairs found on the same web pages.
Thus, our data covers open-ended image relations over a wide distribution, including both visual and beyond-visual ones.

\custompara{Retrieval with Instruction.}
Instruction tuning~\cite{Ouyang2022InstructGPT, Lou2024InstructionSurvey} enables models with strong cross-domain and zero-shot generalization capabilities in retrieving both textual~\cite{Su2023Instructor, Akari2023BERRI_TART} and multimodal content~\cite{Wei2023UniIR}.
However, prior efforts focus on unifying different retrieval tasks with manually-written instructions as task prefixes of actual queries, on a hundred-scale basis.
In contrast, our approach utilizes million-scale instructions that naturally express user's search intents.

\section{\ModelName}
\subsection{Data Construction for Self-Supervised Training}
\label{sec:data-construction}
Web documents contain multimodal contexts, featuring interleaved texts and images on pertinent subjects. Image pairs extracted from the same web page through co-occurrence frequently imply associations between images and specific relations. This encompasses a broad spectrum of image relations, ranging from visual similarity to more nuanced connections (e.g., Figure~\ref{fig:data-construction-overview}). Consequently, these naturally occurring image pairs serve as excellent self-supervised training signals for image retrieval models.
Based on this insight, we propose a systematic data construction pipeline to mine image pairs from web pages and adopt LLMs to generate open-ended instructions that explicitly convey the image relations within each pair.

\custompara{Mining Image Pairs from Web Pages.}
\textit{(1) Grouping \& Cleaning.}
We collect all images with the same URL from Common Crawl\footnote{https://commoncrawl.org/} as a group of images from the same web page for potential pairing.
Due to the inevitable noisy images introduced by simple grouping, we remove duplicated, low-resolution, and advertising images, as well as highly overlapped groups.
This results in a large number of groups with more densely and intrinsically connected images.

\textit{(2) Metadata Expansion.}
To provide detailed textual information of images for later LLMs with massive metadata expansion, we annotate images with Alt-texts\footnote{https://en.wikipedia.org/wiki/Alt\_attribute}, image content annotation (ICA) labels\footnote{https://cloud.google.com/vision/docs/labels}, and captions.
We discard images if their Alt-texts are unqualified.
For ICA labels, we annotate entities for each image, such as general objects and activities.
For image captions, we adopt a SOTA LMM --- PaLI~\cite{chen2023palix} to generate captions.
Each type of metadata provides textual information about images from different perspectives.
See Appendix~\ref{appendix:implementation-details} for more details.

\textit{(3) Scoring \& Filtering.}
After obtaining groups of images along with their extensive metadata, we pair up images within the same group and eliminate unqualified pairs using a combination of relevance measures.
We use the CLIP image-to-image score to assess visual relevance and the text-to-text score for non-visual relevance.
Image pairs that don't meet the criteria, such as those with low scores in both aspects, are excluded from consideration.
To avoid the over-sampling of redundant images and duplicated relations, we set a maximum of three pairs for each group, thereby ensuring a more uniform distribution of images and relations in our training data (see Figure~\ref{fig:word-cloud}).

\custompara{Open-Ended Instruction Generation.}
% \label{subsubsec:instruction-generation}
With the informative metadata of high-quality paired images, LLMs are able to well understand the image content (ICA and caption) and their background information (Alt-text).
Using instruction~\cite{Chung2022FLAN-T5}, few-shot demonstrations~\cite{Brown2020GPT3}, and chain-of-thought prompting~\cite{Wei2022CoT} techniques, PaLM2~\cite{Anil2023PaLM2} generates open-ended instructions that precisely connect the paired images (\textit{image$_q$, image$_t$}).
Figure~\ref{fig:data-construction} illustrates generated instructions and Appendix~\ref{appendix-tab:query-gen} shows the detailed prompt and demonstrations.
Eventually, we obtain 36.7M triplets (\textit{image$_q$, text, image$_t$}) for self-supervised image retrieval training.

\subsection{\ModelName~Model}
\begin{figure}[tb]
    \centering
    % \vspace{-2mm}
    \includegraphics[width=0.96\linewidth]{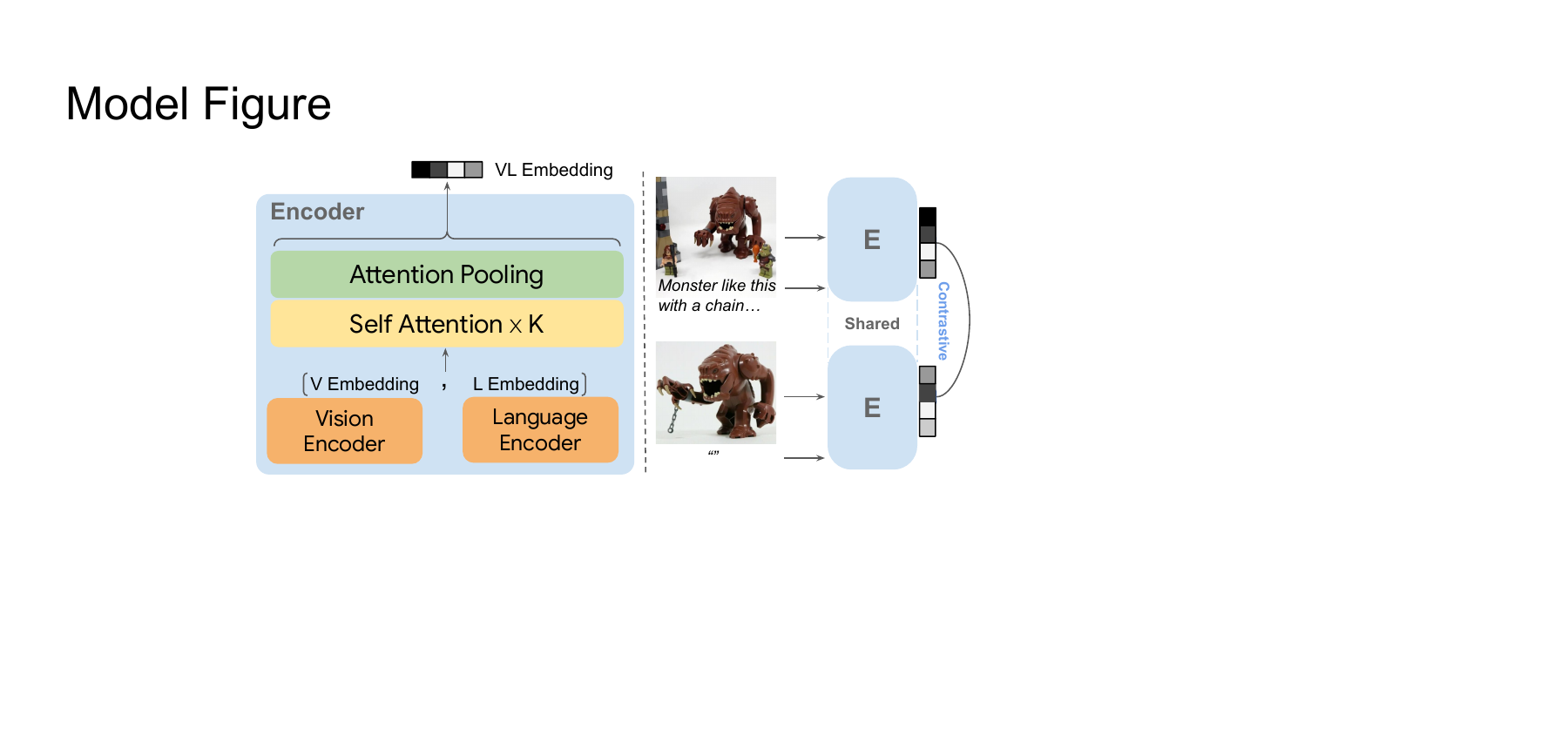}
    \vspace{-2mm}
    \caption{Model architecture and training of \ModelName~Encoder (E), which takes the vision and language embeddings and feeds them as a sequence to self-attention layers for modality integration.}
    \vspace{-4mm}
    \label{fig:model-arch}
\end{figure}

\custompara{Model Design.}
As shown in Figure~\ref{fig:model-arch}, we adopt a simple dual-encoder architecture with shared parameters and initialize the backbone vision and language encoders with CoCa~\cite{Yu2022CoCa} or CLIP~\cite{Radford2021CLIP}.
To enable deep modality integration, we introduce multiple self-attention layers and design a single multi-head attention pooler, compressing the multimodal inputs into a single embedding $\boldsymbol{r}$ for later matching.
Additionally, since the retrieval target comprises only an image without accompanying text, we employ an empty text string \textit{``''} to transform the target into a multimodal input.
We denote $\boldsymbol{r_q}$ as the embedding for the multimodal query (\textit{image$_q$}, \textit{text}) and $\boldsymbol{r_t}$ as the embedding for the target (\textit{image$_t$}, \textit{``''}).
Considering the efficiency, we propose \ModelName-B and \ModelName-L, initialized with the base and large checkpoints, respectively.

\custompara{Model Training.}
We use a simple contrastive loss to train \ModelName. Our model is updated by contrasting the paired query-target against other targets in one training batch.
In particular, as the query image itself (\textit{image$_q$}) can be a challenging hard negative for the multimodal query (\textit{image$_q$, text}), we combine the query image itself and an empty text to encode (\textit{image$_q$}, \textit{``''}) to get $\boldsymbol{r_t^{'}}$ as an additional query negative example. 
To scale up the number of negative examples, for each query image, we use all query negatives and other target negatives in the same batch.
Formally, for the $i$-th training example, the loss function $\mathcal{L}_i$ is defined as,
\begin{equation}
\vspace{-1mm}
    - \text{log} \frac{
                    e^{\text{sim}(\boldsymbol{r_q^i}, \boldsymbol{r_t^i})/\tau}
                }
                {
                \sum_{j=1}^N (
                            e^{\text{sim}(\boldsymbol{r_q^i}, \boldsymbol{r_t^j})/\tau} + e^{\text{sim}(\boldsymbol{r_q^i}, \boldsymbol{r_t^{j'}})/\tau}
                            )
                }
                ,
\nonumber
\vspace{-1mm}
\end{equation}
where $\text{sim}(,)$ indicates a cosine similarity function $\frac{\boldsymbol{r_q}^T\boldsymbol{r_t}}{||\boldsymbol{r_q}||\cdot||\boldsymbol{r_t}||}$, $N$ refers to the sampled batch size, and $\tau$ is a temperature hyperparameter for logit scaling.
Please refer to Appendix~\ref{appendix:implementation-details} for more implementation details.

\section{Experiments}
\subsection{Experiment Setup}
% \usepackage{multirow}
\begin{table*}[t]
\centering
\vspace{-2.5mm}
\small
\caption{Performance comparison on five benchmarks of three multimodality-to-image retrieval tasks. The results of baselines are from the original papers. We mark the best results in bold and the second-best results underlined.
\textsuperscript{$\star$}CIReVL uses multiple model components including ChatGPT for retrieval, we report \# parameters of components with known sizes. 
\textsuperscript{$\dag$}PLI does not release code so we estimate.}
% \begin{tabular}{c|l|cccc|c|c}
\begin{tabular}{l@{}cccccccc}
\toprule

\multicolumn{1}{c}{\multirow{3}{*}{\textbf{Method}}} & \multirow{3}{*}{\textbf{\begin{tabular}[c]{@{}c@{}} Backbone\end{tabular}}} & \multirow{3}{*}{\textbf{\begin{tabular}[c]{@{}c@{}} \# Total \\Params\end{tabular}}} & \multicolumn{4}{c}{\small{\textbf{Composed Image Retrieval}}} & \small{\textbf{Domain Trans}} & \multicolumn{1}{l}{\small{\textbf{Cond}~\textbf{ImSim}}} \\ 
& & & \small{\texttt{{FIQ}}} & \multicolumn{2}{c}{\small{\texttt{CIRR}}} & \small{\texttt{CIRCO}} & \small{\texttt{DTIN}} & \small{\texttt{GeneCIS}} \\
\cmidrule(lr){4-4} \cmidrule(lr){5-6} \cmidrule(lr){7-7} \cmidrule(lr){8-8} \cmidrule(lr){9-9}

% \multirow{2}{*}{\textbf{\begin{tabular}[c]{@{}c@{}}\# Subset\\ Index\end{tabular}}}

& \multicolumn{1}{c}{} & & \textsc{R}@10 & \textsc{R}@1 & \textsc{R}$_s$@1 & m\textsc{AP}@5 & \textsc{R}@10 & \textsc{R}$_s$@1 \\ \midrule
PALAVRA \cite{Cohen2022PALARVA} & \small{\texttt{CLIP-B}} & 176M\pz & 19.8               & 16.6                 & 41.6                 & 4.6                   & -             & -                                    \\
SEARLE \cite{Baldrati2023CIRCO} & \small{\texttt{CLIP-B}} & 165M\pz & 22.9               & 24.0                 & 54.9                 & 9.4                   & -             & -                                    \\
CIReVL \cite{Karthik2023CIReVL} & \small{\texttt{CLIP-B}} & 12.3B\textsuperscript{$\star$} & 28.3                 & 23.9                 & 60.2                 & 14.9                  & -             & \underline{15.9}                                    \\
% % This is using CLIP 
% PLI \cite{Chen2023PLI} & \small{\texttt{ViT-B}} & \textbf{31.3}      & 18.8                 & 44.3                 & 8.1                   & -             & -                                    \\
% This is using BLIP
PLI \cite{Chen2023PLI} & \small{\texttt{BLIP-B}} & 224M\textsuperscript{$\dag$} & \textbf{35.9}      & \underline{27.2}                 & 55.1                 & 7.1                   & -             & -                                    \\
\rowcolor{lightlightgray} \textbf{\ModelName-B} & \small{\texttt{CLIP-B}} & 166M\pz & 26.3 & 27.0        & \underline{66.7}        & \underline{23.1}         & \underline{28.3} & 15.0                        \\ 
\rowcolor{lightgray} \textbf{\ModelName-B} & \small{\texttt{CoCa-B}} & 267M\pz & \underline{35.2} & \textbf{31.6}        & \textbf{69.3}        & \textbf{30.8}         & \textbf{46.8} & \textbf{17.4}                        \\ 
\midrule
Pic2Word \cite{Saito2023Pic2Word} & \small{\texttt{CLIP-L}} & 429M\pz & 24.7 & 23.9                 & -                    & 8.7                   & 10.1          & 11.2                                 \\
SEARLE \cite{Baldrati2023CIRCO} & \small{\texttt{CLIP-L}} & 442M\pz & 25.6               & 24.2                 & 53.8                 & 11.7                  & -             & 12.3                                 \\
Context-I2W \cite{Tang2023ContextI2W} & \small{\texttt{CLIP-L}} & 496M\pz & 27.8               & 25.6                 & -                    & -                     & 12.9          & -                                    \\
CompoDiff \cite{Gu2023CompoDiff} & \small{\texttt{CLIP-L}} & 568M\pz & \underline{36.0}      & 18.2                 & 57.4                 & 12.6                  & -             & 14.9                                 \\
CIReVL \cite{Karthik2023CIReVL} & \small{\texttt{CLIP-L}} & 12.5B\textsuperscript{$\star$} & 28.6                & 24.6                 & 59.5                 & 18.6                  & -             & 15.9                                    \\
PLI \cite{Chen2023PLI} & \small{\texttt{CLIP-L}} & 428M\textsuperscript{$\dag$} & 35.4               & 25.5                 & 55.6                 & 10.4                  & -             & -                                    \\
LinCIR \cite{Gu2023LinCIR} & \small{\texttt{CLIP-L}} & 442M\pz & 26.3 & 25.0 & 57.1 & 12.6 & - & 12.2 \\
\rowcolor{lightlightgray} \textbf{\ModelName-L} & \small{\texttt{CLIP-L}} & 465M\pz & 30.7 & \underline{30.1}        & \underline{68.1}        & \underline{29.6}         & \underline{41.5} & \underline{16.3}                        \\ 
\rowcolor{lightgray} \textbf{\ModelName-L} & \small{\texttt{CoCa-L}} & 613M\pz & \textbf{38.0} & \textbf{33.3} & \textbf{70.9} & \textbf{34.1} & \textbf{48.2} & \textbf{16.7}                        \\ 
\bottomrule
\vspace{-7.5mm}
\end{tabular}
\label{tab:main-res}
\end{table*}

\custompara{Benchmarks and Metrics.}
To comprehensively evaluate \ModelName' multimodality-to-image retrieval ability, we consider three related tasks in a zero-shot, one-checkpoint setting: (1) composed image retrieval (CIR), (2) domain transfer retrieval, and (3) conditional image similarity.
Each task has different yet limited sets of image relations.
Table~\ref{tab:dataset-statistics} shows the detailed statistics of five benchmarks.
\begin{table}[t]
\vspace{-2mm}
\centering
\caption{Statistics of five evaluation benchmarks.
We average the number of queries over sub-tasks (e.g., FIQ), if available.
The \# Index represents the size of the retrieval pool that is shared amongst all queries and the \# Subset Index is the average size of subsets, each of which is dedicated for a query.
}
\resizebox{0.96\linewidth}{!}{

\begin{tabular}{@{\;}l@{\;}c@{\;}cccc@{\;}}
\toprule
& \multicolumn{2}{c}{\textbf{Open-Domain}} & \multirow{2}{*}{\textbf{\# Query}} & \multirow{2}{*}{\textbf{\# Index}} & \multirow{2}{*}{\textbf{\begin{tabular}[c]{@{}c@{}}\# Subset\\ Index\end{tabular}}} \\
& \texttt{Image?}   & \texttt{Instr?}  & & & \\ \midrule
\texttt{FIQ}     & \xmark           & \xmark                & 2,005                             & 5,179                             & -                                                                                  \\
\texttt{CIRR}    & \cmark           & \xmark                & 4,148                             & 2,316                             & 8.3                                                                                \\
\texttt{CIRCO}   & \cmark           & \xmark                & 800                               & 123,403                           & -                                                                                  \\
\texttt{DTIN}      & \cmark           & \xmark                & 10,000                            & 16,983                            & -                                                                                  \\
\texttt{GeneCIS} & \cmark           & \xmark                & 2,008                             & -                                 & 13.8  % \\ \midrule
% Ours             & \cmark           & \cmark                & 100                               & 1,360,561                         & -

\\ \bottomrule    
\vspace{-10.5mm}
\end{tabular}
}

\label{tab:dataset-statistics}
\end{table}

\textbf{Composed Image Retrieval.} We consider one domain-specific and two open-domain benchmarks to evaluate the model's domain adaptability and its capability on real-world natural images, respectively.
\texttt{FIQ}~\cite{Wu2021FashionIQ} is a fashion-domain benchmark with three disjoint retrieval sub-tasks: dress, shirt, and toptee.
Following previous work~\cite{Saito2023Pic2Word, Baldrati2023CIRCO, Gu2023LinCIR}, we evaluate on its validation set and report recall averaged over sub-tasks.
\texttt{CIRR}~\cite{Liu2021CIRR} is the first dataset constructed on natural images~\cite{Suhr2019NLVR2} with nine pre-defined relations between the query and the target images.
It also includes a subset retrieval setting where models retrieve a target image from a dedicated small subset for each query.
However, in addition to the limited size of the retrieval pool, it also suffers from false negative issues, as pointed out by~\citet{Baldrati2023CIRCO}.
We utilize recall (\textsc{R} and \textsc{R}$_s$) to evaluate standard retrieval and subset retrieval.
In contrast, to better align with the real-world large-scale retrieval, \texttt{CIRCO}  annotates multiple ground truths for each query and has over 120K natural images~\cite{Lin2014MSCOCO} as the index set.
Therefore, we regard CIRCO as our main benchmark.
As each query has multiple targets, we adopt mean Average Precision (mAP) as the evaluation metric.

\textbf{Domain Transfer Retrieval.} Domain Transfer ImageNet (\texttt{DTIN};~\citet{Saito2023Pic2Word}) aims to retrieve an image from another domain with the same conceptual object shown in the query image.
It is constructed from natural images in ImageNet~\cite{Deng2009ImageNet} and images in other domains in ImageNet-R~\cite{Hendrycks2021ImageNetR}.
For example, given a domain keyword \textit{``cartoon''} and a real horse image as a query, models are expected to retrieve a cartoon horse from the index set with images from multiple domains.
It covers 4 domains, 10K objects, and over 16K images as the index set.
Following prior works~\cite{Saito2023Pic2Word, Karthik2023CIReVL}, we report recall averaged over sub-tasks.

\textbf{Conditional Image Similarity.} \texttt{GeneCIS}~\cite{Vaze2023GeneCIS} is a keyword-conditioned image similarity measurement benchmark.
It has four sub-tasks about changing or focusing the attribute or object in the given image.
For each query image and keyword, models need to find the most similar images to the query image, conditioned on the given keyword, from a dedicated small subset with 13.8 images on average.
For example, in the change-object sub-task with keyword \textit{``car''} and an image, models need to find another image depicting a similar scene but with additional cars.

\custompara{Baselines.}
We consider several baselines: (1) PALARVA~\cite{Cohen2022PALARVA}, (2) Pic2Word~\cite{Saito2023Pic2Word}, (3) SEARLE~\cite{Baldrati2023CIRCO}, (4) ContextI2W~\cite{Tang2023ContextI2W}, (5) LinCIR~\cite{Gu2023LinCIR}, (6) CIReVL~\cite{Karthik2023CIReVL} (7) CompoDiff~\cite{Gu2023CompoDiff} and (8) PLI~\cite{Chen2023PLI}.
Details of these methods are described in Appendix~\ref{appendix:baselines}.

% \subsection{Main Results}
\subsection{Multimodality-to-Image Retrieval}
\label{sec:multimodality-to-image-retrieval}
\begin{table*}[t]
\centering
\small
\vspace{-1em}
\caption{\label{tab:zs-it-rt} Zero-shot image-text retrieval results. Results are marked in bold if they are better than initialized checkpoints. \textsuperscript{$\star$}CoCa we reproduced and used for \ModelName. \textsuperscript{$\dagger$}CoCa reported in the original paper.}
% \resizebox{1.0\textwidth}{!}{%
% \begin{tabular}{@{}lcccccccccccc@{}}
\begin{tabular}{lcccccccccccc}
    \toprule 
    \multicolumn{1}{c}{\multirow{3}{*}{\textbf{Model}}} & \multicolumn{6}{c}{\texttt{Flickr30K} (1K test set)} & \multicolumn{6}{c}{\texttt{MSCOCO} (5K test set)} \\
    & \multicolumn{3}{c}{Image $\rightarrow$ Text} & \multicolumn{3}{c}{Text $\rightarrow$ Image} & \multicolumn{3}{c}{Image $\rightarrow$ Text} & \multicolumn{3}{c}{Text $\rightarrow$ Image} \\
    \cmidrule(lr){2-4} \cmidrule(lr){5-7} \cmidrule(lr){8-10} \cmidrule(lr){11-13}
     & R@1 & R@5 & R@10 & R@1 & R@5 & R@10 & R@1 & R@5 & R@10 & R@1 & R@5 & R@10 \\
    \midrule
    % CoCa-Base & 89.8 & 98.8 & 99.8 & 76.8 & 93.7 & 96.8 & 63.8 & 84.7 & 90.7 & 47.5 & 72.4 & 80.9 \\
    \textcolor{gray}{CoCa-B\textsuperscript{$\dagger$}} & \textcolor{gray}{89.8} & \textcolor{gray}{98.8} & \textcolor{gray}{99.8} & \textcolor{gray}{76.8} & \textcolor{gray}{93.7} & \textcolor{gray}{96.8} & \textcolor{gray}{63.8} & \textcolor{gray}{84.7} & \textcolor{gray}{90.7} & \textcolor{gray}{47.5} & \textcolor{gray}{72.4} & \textcolor{gray}{80.9} \\
    CoCa-B\textsuperscript{$\star$} & 88.6 & 98.5 & 99.4 & 74.5 & 93.4 & 96.4 & 63.4 & 84.2 & 90.4 & 46.4 & 71.5 & 80.1 \\
    \rowcolor{lightgray} \textbf{\ModelName-B} & 87.9 & 97.7 & \textbf{99.5} & \textbf{76.2} & \textbf{93.7} & \textbf{96.5} & \textbf{64.8} & \textbf{85.5} & \textbf{91.2} & \textbf{48.9} & \textbf{73.9} & \textbf{82.5} \\ \midrule
    % CoCa-Large & 91.4 & 99.2 & 99.9 & 79.0 & 95.1 & 97.4 & 65.4 & 85.6 & 91.4 & 50.1 & 73.8 & 81.8 \\
    \textcolor{gray}{CoCa-L\textsuperscript{$\dagger$}} & \textcolor{gray}{91.4} & \textcolor{gray}{99.2} & \textcolor{gray}{99.9} & \textcolor{gray}{79.0} & \textcolor{gray}{95.1} & \textcolor{gray}{97.4} & \textcolor{gray}{65.4} & \textcolor{gray}{85.6} & \textcolor{gray}{91.4} & \textcolor{gray}{50.1} & \textcolor{gray}{73.8} & \textcolor{gray}{81.8} \\
    CoCa-L\textsuperscript{$\star$} & 92.1 & 98.8 & 99.9 & 78.4 & 94.2 & 96.9 & 65.1 & 85.5 & 91.3 & 49.3 & 73.2 & 81.5\\
    \rowcolor{lightgray} \textbf{\ModelName-L} & 89.6 & 98.7 & 99.4 & \textbf{79.7} & \textbf{95.0} & \textbf{97.4} & \textbf{67.7} & \textbf{87.6} & \textbf{92.7} & \textbf{53.1} & \textbf{77.4} & \textbf{84.9}\\
    \bottomrule

\end{tabular}
% }

\vspace{-1.5em}
\end{table*}
Table~\ref{tab:main-res} shows results over five benchmarks from three tasks, from which we have the following observations:

First, with the comparable model size, both CLIP- and CoCa-based \ModelName outperform previous state-of-the-art models across the four open-domain benchmarks by a large margin, especially CoCa-based \ModelName-L on the challenging CIRCO (mAP@5 from 12.6 to 34.1) and DTIN (R@10 from 12.9 to 48.2).
This shows the strong capability of \ModelName.
We leave full results on Appendix~\ref{appendix:detailed-comparison-results} and detailed parameter efficiency analysis on \S~\ref{sec:model-analysis}.

Second, by comparing \ModelName-L to \ModelName-B, we find generally consistent performance improvements across five benchmarks.
This demonstrates the constructed data is of high quality and can benefit larger models.
Also, this observation shows the scalability thanks to the simple dual-encoder model architecture and contrastive loss.

\subsection{Image-to-Image Retrieval}
\label{sec:image-to-image-retrieval}
\begin{table}[t]
\centering
\vspace{-2mm}
% \small
\caption{Results on three image-to-image retrieval benchmarks. The results of baselines are from~\citet{Lin2023ZSESBIR} with separate checkpoints for each benchmark while \ModelName (ML) models are evaluated across benchmarks under one-checkpoint setting.}
\resizebox{1.0\linewidth}{!}{
\begin{tabular}{lcccccc}
\toprule
\multicolumn{1}{c}{\multirow{2}{*}{\textbf{Method}}} & \multicolumn{2}{c}{\texttt{TU-Berlin}} & \multicolumn{2}{c}{\texttt{Sketchy}} & \multicolumn{2}{c}{\texttt{QuickDraw}} \\  \cmidrule(lr){2-3} \cmidrule(lr){4-5} \cmidrule(lr){6-7}
\multicolumn{1}{c}{}                                 & mAP                & P@100             & mAP@200           & P@200            & mAP                & P@200             \\ \midrule
ViT~(\citeyear{Dosovitskiy2021ViT})                        & 36.0               & 50.3              & 40.3              & 51.2             & 10.1               & 11.3              \\
SOTA~(\citeyear{Lin2023ZSESBIR})                     & 56.9               & 63.7              & 52.5              & 62.4             & 14.5               & 21.6              \\ \midrule
\rowcolor{lightgray} \textbf{ML-B}-CLIP           & 45.9               & 57.9              & 49.3              & 60.6             & 10.1               & 14.0              \\
\rowcolor{lightgray} \textbf{ML-B}-CoCa           & 61.7               & 72.1              & 70.5              & 77.2             & 13.9               & 19.9              \\
\rowcolor{lightgray} \textbf{ML-L}-CLIP           & 62.9               & 73.1              & 68.2              & 75.8             & 15.1               & 20.4              \\
\rowcolor{lightgray} \textbf{ML-L}-CoCa           & \textbf{70.2}      & \textbf{79.1}     & \textbf{75.7}     & \textbf{81.3}    & \textbf{19.7}      & \textbf{27.4}    \\ \bottomrule
\end{tabular}
}
\vspace{-10.5mm}
\label{tab:sbir-res}
\end{table}
Although \ModelName models are trained for (\textit{image$_q$}, \textit{text}) $\rightarrow$ \textit{image$_t$} task format, they can naturally cover \textit{image$_q$} $\rightarrow$ \textit{image$_t$} tasks by providing a fixed text instruction for all \textit{image$_q$}.
As a case study, we consider zero-shot sketch based image retrieval (ZS-SBIR) task where models need to retrieve a natural image given a sketch of it.
By simply using \textit{``find a natural image of it''} for all query images, \ModelName can perform such a task.

Following the prior zero-shot SOTA methods~\cite{Liu2019ZSSBIR, Lin2023ZSESBIR} in this domain, we consider three benchmarks, namely TU-Berlin~\cite{Zhang2016TUBerlin}, Sketchy~\cite{Yelamarthi2018Sketchy}, and QuickDraw~\cite{Dey2019QuickDraw}.
TU-Berlin has 30 classes, 2,400 sketch queries, and 27,989 natural images as index set;
Sketchy consists of 21 classes unseen in ImageNet-1K and 12,694 queries over an index set with 12,553 natural images;
QuickDraw has 30 classes, 92,291 queries, and a 54,146-sized index set.
For each dataset, we report mAP and precision metrics used in the prior SOTA work~\cite{Lin2023ZSESBIR}.

Notably, unlike previous zero-shot methods that use separate checkpoints trained on each dataset and evaluated on the above holdout test set, we use the same checkpoints for evaluation on all benchmarks.
Results are reported in Table~\ref{tab:sbir-res} and we can find that our models outperform prior SOTA methods by a significant margin, despite our adherence to a single checkpoint setting.
This demonstrates the strong generalization capability of \ModelName models and the diversity of tasks that they can cover.

\subsection{Text-to-Image Retrieval}
\label{sec:text-to-image-retrieval}
Since \ModelName models are built upon vision and language encoders, these backbone encoders after training can still be reused for \textit{image} $\rightarrow$ \textit{text} and \textit{text} $\rightarrow$ \textit{image} retrieval tasks.
Therefore, we evaluate \ModelName' backbone encoders on Flickr30k~\cite{Plummer2015Flickr30k} and MSCOCO~\cite{chen2015MSCOCOCaptions}, using the same dataset splits and evaluation metrics as prior works~\cite{Radford2021CLIP, Yu2022CoCa}.

Table~\ref{tab:zs-it-rt} shows the comparison between the original encoders and the ones updated after \ModelName training.
For \textit{text} $\rightarrow$ \textit{image} task, we can observe consistent and non-trivial improvements across all metrics on both datasets.
For \textit{image} $\rightarrow$ \textit{text} task, we observe marginal drops.
These observations show that our training recipe can enhance the backbone encoders for text-to-image retrieval.
We can draw the same conclusion with CLIP, which is detailed in Table~\ref{appendix-tab:clip-zs-it-rt}.
The improvements might stem from the fact that our multimodality-to-image training task necessitates deep understanding of text instruction, thus improving backbone language encoders.
These text-to-image results, along with the results on image-to-image and multimodality-to-image tasks, show that \ModelName can well handle various forms of image retrieval tasks, all with strong performances.

\section{Analysis}
\subsection{Data Analysis}
\label{sec:data-analysis}

\custompara{Comparison to Existing Training Data.}
Previous data construction efforts, including CompoDiff~\cite{Gu2023CompoDiff} and InstructPix2Pix (IP2P;~\citet{Brooks2023InstructPix2Pix}), use synthesized image pairs and essentially template-based instructions to train image retrieval models.
Given the data availability and the fact that CompoDiff adopts a creation pipeline similar to IP2P, we use IP2P data as our baseline to explore the effects of different training data on downstream models.
We compare a CoCa-based \ModelName-B model trained on all IP2P data (1M) with one trained on our down-sampled, same-sized data, using the same training recipe.
Table~\ref{tab:ip2p-training-comparison} shows that \ModelName + Ours achieves performance advantages over its variant trained with IP2P data (\ModelName + IP2P) on all five benchmarks. 
This proves that our data with natural images and template-free instructions can enable stronger image retrieval models.
Please refer to Appendix~\ref{appendix:detailed-comparison-results} for detailed comparisons.

In addition, we compare these two models with IP2P-trained CompoDiff, which is a retrieval model designed for using synthesized images.
Despite its specific design, \ModelName + IP2P still outperforms the CompoDiff + IP2P.
Also, it achieves better results on CIRCO, DTIN, and GeneCIS than prior comparable-sized SOTA baselines.
These show the advantage of our training recipe, as our model can achieve decent results even when trained on sub-optimal data.

\begin{table}[t]
\small
\centering
% \vspace{-2.5mm}
\caption{Performance comparison of CoCa-based \ModelName-B trained with 1M IP2P data against 1M our data.
We report averaged R@10 \& R@50 on FIQ and averaged R@1 \& R$_s$@1 CIRR for comparisons with CompoDiff~\cite{Gu2023CompoDiff}.}

{
\begin{tabular}{@{}l@{\quad}c@{\;\;}c@{\;\;}c@{\;\;}c@{\;\;}c@{}}
\toprule
& \texttt{FIQ} & \texttt{CIRR} & \texttt{CIRCO} & \texttt{DTIN} & \texttt{GeneCIS} \\
& {\tiny (R$_{\texttt{AVG}}$)} & {\tiny (R$_{\texttt{AVG}}$)} & {\tiny (mAP@5)} & {\tiny ~(R@10)~} & {\tiny (R@1)} \\ 
\midrule
{\footnotesize CompoDiff + IP2P} & 27.2                                   & 27.4                                           & -              & -             & - \\ \midrule
{\footnotesize \ModelName + IP2P}   & 29.8                                & 33.7                                           & 13.6            & 30.2          & 14.5  \\
{\footnotesize \ModelName + Ours} & \textbf{43.7}     & \textbf{48.2}                                  & \textbf{29.7}  & \textbf{43.7} & \textbf{15.8} \\ \bottomrule
\end{tabular}
}
% \vspace{-2mm}
\label{tab:ip2p-training-comparison}
\end{table}

\begin{figure}[t]
\vspace{-2.5mm}
    \centering
    \subfigure[IP2P]{\label{fig:ip2p-wordcloud}\includegraphics[width=0.48\linewidth]{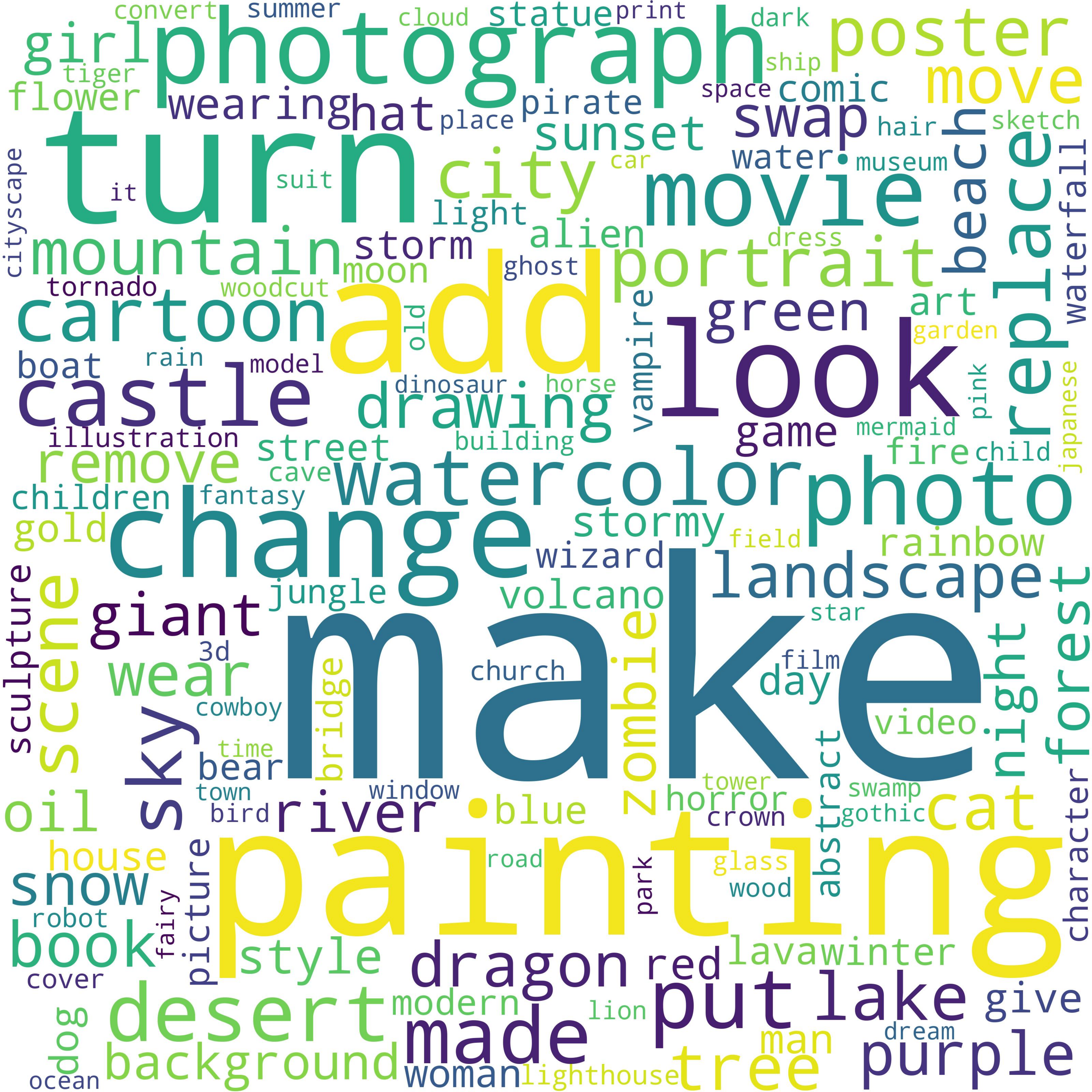}}
    \subfigure[Our training data]{\label{fig:ours-wordcloud}\includegraphics[height=0.48\linewidth]{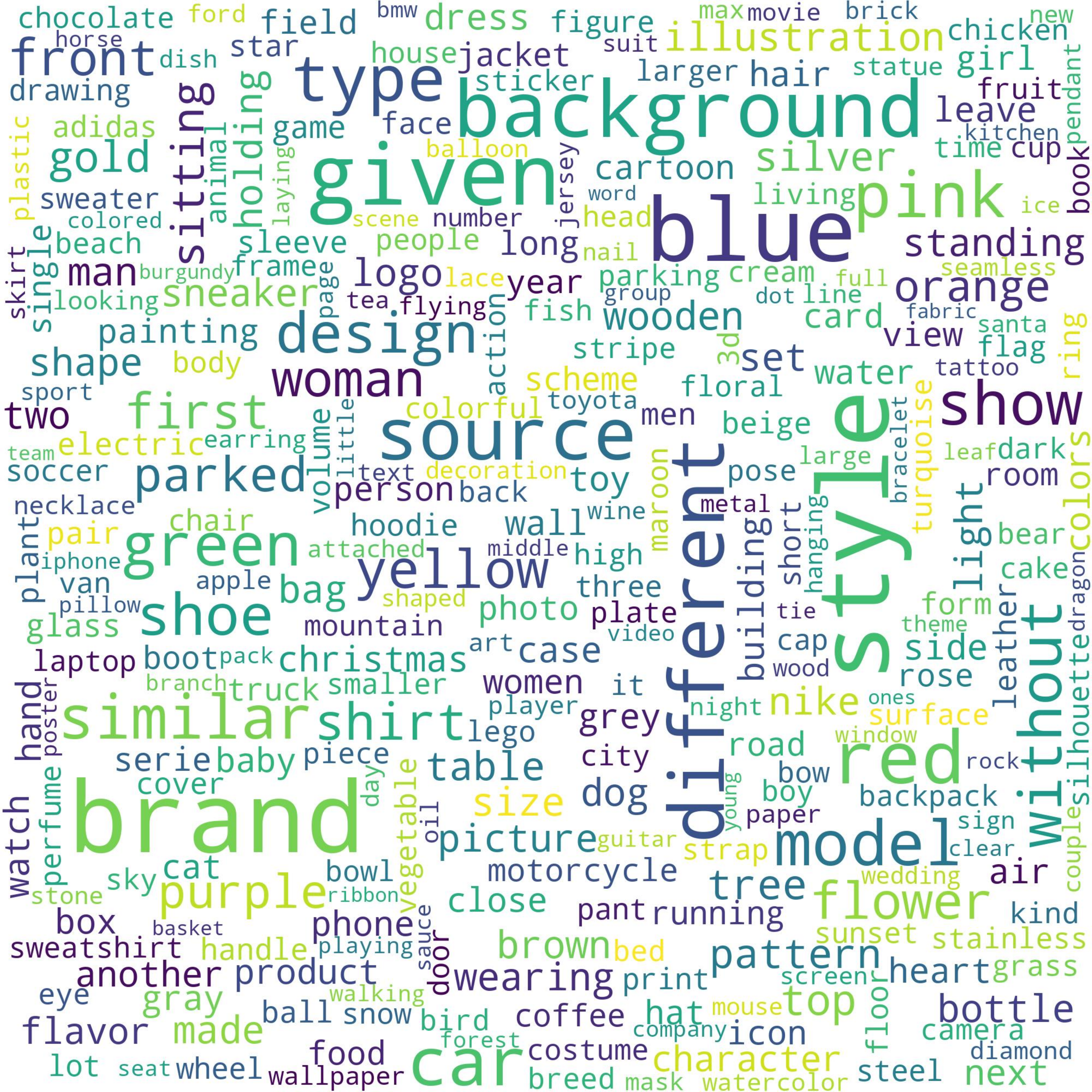}}
    \vspace{-2mm}
    \caption{Word distributions of IP2P data and our data.}
    \vspace{-5mm}
    \label{fig:word-cloud}
\end{figure}

\begin{figure}[t]
% \centering
\pgfplotsset{compat=1.15}

\begin{tikzpicture}
\begin{axis}[
    width=\linewidth,
    height=0.8\linewidth,
    xlabel={\# Our Training Data (M)},
    ylabel={Performance},
    xtick=data,
    xmode=log,    %<- here
    log basis x={10},
    xticklabels={0.2, 0.5, 1, 5, 10, 25, 37},
    scaled x ticks=false,
    % legend pos=north west,
    ymajorgrids=true,
    grid style=dashed,
    legend style={
    at={(0.15,1)},
    anchor=north,
    font=\small,
    text opacity=0.7,
    fill opacity=0.3,
    draw=none},
    legend cell align={left},
]

% FashionIQ (R@10)
\addplot[
    color=blue,
    mark=star,
    opacity=0.4,
    ]
    coordinates {
    (200000, 28.3)(500000, 31.3)(1080852,33.5)(5433310,33.7)(10867520,35.1)(25824678,35.0)(36714118,35.2)
    };
    % \addlegendentry{FIQ (R@10)}
    \addlegendentry{FIQ}

% CIRR (R@1)
\addplot[
    color=orange,
    mark=o,
    opacity=0.4,
    ]
    coordinates {
    (200000, 21.7)(500000, 24.7)(1080852,29.6)(5433310,29.8)(10867520,31.7)(25824678,32.0)(36714118,31.6)
    };
    % \addlegendentry{CIRR (R@1)}
    \addlegendentry{CIRR}

% CIRCO (mAP@5)
\addplot[
    color=green,
    mark=square,
    opacity=0.4,
    ]
    coordinates {
    (200000, 22.4)(500000, 25.7)(1080852,29.7)(5433310,30.4)(10867520,30.5)(25824678,30.3)(36714118,30.8)
    };
    % \addlegendentry{CIRCO (mAP@5)}
    \addlegendentry{CIRCO}

% DomainTrans (R@10)
\addplot[
    color=red,
    mark=x,
    opacity=0.4,
    ]
    coordinates {
    (200000, 43.1)(500000, 46.0)(1080852,43.7)(5433310,46.2)(10867520,46.7)(25824678,47.4)(36714118,46.8)
    };
    % \addlegendentry{DT (R@10)}
    \addlegendentry{DTIN}

% GeneCIS (R@1)
\addplot[
    color=black,
    mark=triangle,
    opacity=0.4,
    ]
    coordinates {
    (200000, 14.8)(500000, 16.1)(1080852,15.8)(5433310,15.9)(10867520,16.3)(25824678,16.4)(36714118,17.4)
    };
    % \addlegendentry{GeneCIS (R@1)}
    \addlegendentry{GeneCIS}

% Average Performance
\addplot[line width=0.75pt, color=myRed,mark=diamond*, mark options={scale=1.25, fill=myRed}]
    coordinates {
    (200000, 26.1)(500000, 28.7)(1080852,30.5)(5433310,31.2)(10867520,32.1)(25824678,32.2)(36714118,32.4)
    };
    \addlegendentry{\textbf{Average}}

\end{axis}
\end{tikzpicture}
\vspace{-0.8em}
% \end{document}
\vspace{-2.5mm}
\caption{Performance of CoCa-based \ModelName-B when trained with different sizes of our data.}
\label{fig:data-scaling}
\vspace{-5mm}
\end{figure}
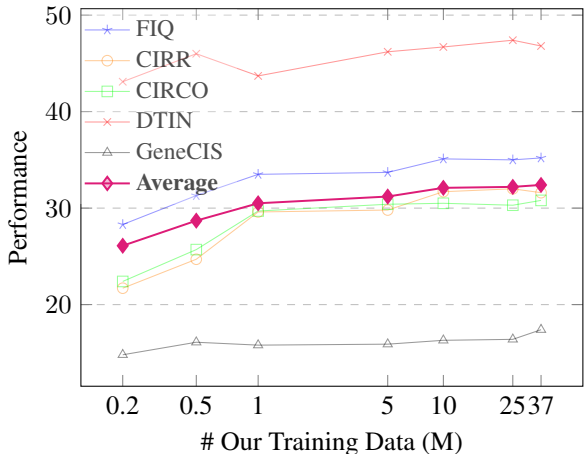

To provide more insights, we visualize the words of instructions in IP2P and in our data separately in Figure~\ref{fig:word-cloud}.
As we can see, IP2P data has a large number of instruction keywords like \textit{``turn''} and \textit{``make''} due to its template-based nature.
Also, it has many coarse-grained keywords such as \textit{``photograph''} and \textit{``painting''}.
In contrast, because of the controlled sampling from one web page described in \S~\ref{sec:data-construction}, our data has more diverse and equally distributed keywords, covering fine-grained labels like \textit{``brand''}.

\custompara{Data Scaling.}
To investigate the effect of our data scale on models, we train CoCa-based \ModelName-B on randomly sampled sets of 0.2M, 0.5M, 1M, 5M, 10M, 25M, and the entire 36.7M triplets.
Results on five benchmarks and their average performance are illustrated in Figure~\ref{fig:data-scaling}.
As the data size increases, \ModelName shows enhanced average performance, especially before the 10M mark.
This indicates the effectiveness of scaling data.

\custompara{Impacts of Instructions during Training.}
Instructions used in previous works~\cite{Brooks2023InstructPix2Pix, Gu2023CompoDiff} are rooted from templates while our instructions are template-free.
To investigate the effects of different instructions on downstream models, we also synthesize template-based instructions for naturally occurring image pairs collected in \S~\ref{sec:data-construction}.
Specifically, due to the massive informative metadata of each image, we utilize LLMs to determine key metadata to fill pre-defined sentence structures.
For our template-free instructions, LLMs are specifically guided to generate diverse and coherent instructions without adhering to any fixed template.
We show concrete examples of different instructions in Figure~\ref{appendix-fig:instruction-comparison} in the Appendix.

\begin{table}[t]
\small
% \vspace{-2mm}
\centering
\caption{Results of CoCa-based \ModelName-B trained with template-based and template-free instructions, at 1M scale.}
\label{tab:instruction-comparison}
\begin{tabular}{l@{\;\;}c@{\;\;}c@{\;\;}c@{\;\;}c@{\;\;}c}
\toprule
\multicolumn{1}{c}{\multirow{2}{*}{\textbf{Instruction}}}  & \texttt{FIQ} & \texttt{CIRR} & \texttt{CIRCO} & \texttt{DTIN} & \texttt{GeneCIS} \\
 & R@10               & R@1           & mAP@5          & R@10                 & R@1              \\ \midrule
\cellcolor[HTML]{D9D2E9}Template-based & 33.4               & 23.4          & 25.1           & 23.1                 & 14.6             \\
\cellcolor[HTML]{C9DAF8}Template-free  & \textbf{33.5}               & \textbf{29.6}          & \textbf{29.7}           & \textbf{43.7}                 & \textbf{15.8}             \\ \bottomrule
\end{tabular}
% }
\vspace{-4.5mm}

\end{table}
Table~\ref{tab:instruction-comparison} compares the performance of two CoCa-based \ModelName-B models.
Both of them are trained on 1M triplets, using the same image pairs but different instructions mentioned above.
Template-free instructions clearly result in a stronger model, as evidenced by consistently better results on all benchmarks compared to the other model.
This demonstrates that naturally expressed and diverse instructions can better stimulate the model to understand image relations and follow instructions.

\begin{figure*}[tbh]
\centering
\usepgfplotslibrary{groupplots}
\pgfplotsset{compat=1.15}
\centering
\begin{tikzpicture}
  \begin{groupplot}[
    % xmin=1.8e8,
    % xmax=3.1e9,
    xmin=2e8,
    xmax=1.8e10,
    group style={
      group size=3 by 1,
      xlabels at=edge bottom,
      ylabels at=edge left,
      horizontal sep=1cm,
    },
    scaled x ticks=false,
    width=6.2cm,
    height=5cm,
    xlabel={\# Parameters (B)},
    ylabel={Performance},
    xmode=log,
    log basis x={10},
    xtick={2e8, 2e9, 1.5e10},
    xticklabels={0.2, 2, 15},
    ymajorgrids=true,
    grid style=dashed,
    legend style={at={(-0.7,1.35)},anchor=north, legend columns=-1}
    ]

    % Second subplot: CIRCO
    \nextgroupplot[title=CIRCO (mAP@5),
    legend to name=grouplegend,
    title style={yshift=-1.5ex},
    ytick={6, 10, 14, 18, 22, 26, 30, 34}
    ]
    \addplot[line width=0.75pt, color=myBlue,mark=diamond*, mark options={scale=1.25}] coordinates {(267459585, 30.8) (613155841, 34.1)};
    \addplot[color=black,mark=*] coordinates {(428666880, 8.72) (987422208, 11.65) (2541142272, 5.54)};
    \addplot[color=myGreen,mark=square*] coordinates {(441779200, 11.68) (1011284480, 16.08) (2553729792, 13.2)};
    \addplot[line width=0.75pt,color=brown,mark=x] coordinates {(441779200, 12.59) (1011284480, 17.6) (2578900224, 19.71)};
    \addplot[color=magenta,mark=triangle*] coordinates {(567660304, 12.55) (2877173264, 15.33)};
    \addplot[color=orange,mark=pentagon*] coordinates {(12250000000, 14.94) (12530000000, 18.57) (14600000000, 26.77)};

    % Third subplot: DTIN
    \nextgroupplot[
    title=DTIN (R@10), 
    title style={yshift=-1.5ex},
    ytick={10,20,30,40,50,60}
    ]
    \addplot[line width=0.75pt, color=myBlue,mark=diamond*, mark options={scale=1.25}] coordinates {(267459585, 46.8) (613155841, 48.2)};
    \addplot[color=black,mark=*] coordinates {(428666880, 10.1)};
    % \addplot[color=green,mark=square,] coordinates {(nan, nan)};
    % \addplot[color=brown,mark=x,] coordinates {(nan, nan)};
    % \addplot[color=magenta,mark=triangle,] coordinates {(nan, nan)};
    \addplot[color=orange,mark=pentagon*] coordinates {(14600000000, 23.8)};
    
    % First subplot: GeneCIS
    \nextgroupplot[title=GeneCIS (R@1),
    title style={yshift=-1.5ex},
    ytick={11,12,13,14,15,16,17}
    ]
    \addplot[line width=0.75pt, color=myBlue,mark=diamond*, mark options={scale=1.25}] coordinates {(267459585, 17.4) (613155841, 16.7)};
    \addplot[color=black,mark=*] coordinates {(428666880, 11.16) (987422208, 11.89) (2541142272, 10.67)};
    \addplot[color=myGreen,mark=square*] coordinates {(441779200, 12.26) (1011284480, 13.34) (2553729792, 12.87)};
    \addplot[line width=0.75pt,color=brown,mark=x] coordinates {(441779200, 12.19) (1011284480, 13.76) (2578900224, 13.66)};
    \addplot[color=magenta,mark=triangle*] coordinates {(567660304, 14.88) (2877173264, 15.48)};
    \addplot[color=orange,mark=pentagon*] coordinates {(12250000000, 15.9) (12530000000, 15.9) (14600000000, 17.4)};
    \legend{\ModelName~~~~, Pic2Word~~~~, SEARLE~~~~, LinCIR~~~~, CompoDiff~~~~, CIReVL~~~~}

  \end{groupplot}

  \node at ($(group c1r1.south)!0.5!(group c3r1.south)+(0,-0.7cm)$) [inner sep=0pt,anchor=north] {\ref{grouplegend}};

\end{tikzpicture}
\vspace{-6.5mm}
\caption{
    Model Size vs. Performance. \ModelName-B outperforms the SOTA CIReVL on three tasks even with 50$\times$ smaller \# Parameters.
    }
    \vspace{-4.5mm}
\label{fig:params-efficiency}
\end{figure*}
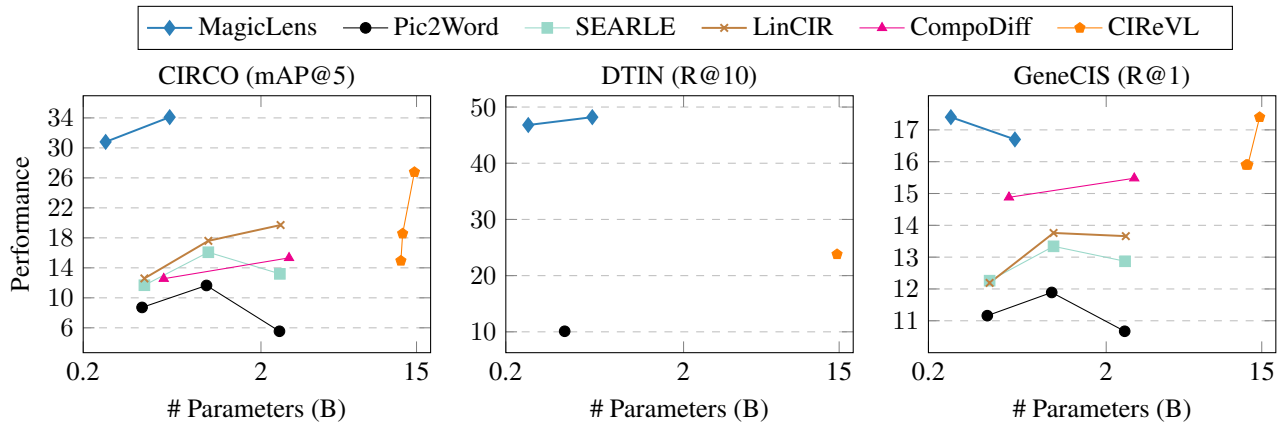

\subsection{Model Analysis}
\label{sec:model-analysis}
\custompara{Model Size vs. Performance.}
Previous SOTA methods~\cite{Gu2023LinCIR, Gu2023CompoDiff} consider using both larger vision and language encoders~\cite{Cherti2023OpenCLIP} or using LMMs and LLMs on-the-fly~\cite{Karthik2023CIReVL} for performance benefits.
However, we argue that the model sizes and the correlated efficiencies should also be taken into consideration for real-world deployments.
In Figure~\ref{fig:params-efficiency}, we visualize the relationship between model size and performance of various models on GeneCIS, CIRCO, and DTIN benchmarks. The results on GeneCIS and CIRCO are from~\citet{Gu2023LinCIR, Gu2023CompoDiff} and using CLIP-Large, OpenCLIP-Huge, and OpenCLIP-Giant backbones~\cite{Radford2021CLIP, Cherti2023OpenCLIP}.
Results of CIReVL~\cite{Karthik2023CIReVL} on DTIN are not fully reported by the authors.
We omit the size of ChatGPT used and only count parameters of CIReVL's other model components (e.g., BLIP2-FLANT5-XXL + OpenCLIP-Giant).

Despite the 50$\times$ smaller size of CoCa-based \ModelName-B (267M) compared to other baselines (e.g., CIReVL with 14.6B), it achieves better performance on these benchmarks, with a significant advantage on the DTIN.
This observation demonstrates the high parameter efficiency introduced by the parameter-sharing design in our model and the strong advantage of our data in enabling strong yet small models.
Detailed results are in Table~\ref{appendix-tab:full-circo},~\ref{appendix-tab:full-dt}, and~\ref{appendix-tab:full-cond-img-sim} in Appendix.

\begin{table}[t]
\small
\centering
% \resizebox{0.96\linewidth}{!}
\vspace{-2.5mm}
\caption{Ablation study on CoCa-based \ModelName-B taking query images as negative samples during training (Qry Neg).}
{
\begin{tabular}{@{\;}l@{\;\;\;}c@{\;\;}c@{\;\;}c@{\;\;}c@{\;\;}c@{\;}}
\toprule
& \texttt{FIQ} & \texttt{CIRR} & \texttt{CIRCO} & \texttt{DTIN} & \texttt{GeneCIS} \\
& R@10               & R@1           & mAP@5          & R@10                             & R@1                         \\ \midrule
\ModelName & 35.2               & 31.6          & 30.8           & 46.8                                      & 17.4 \\
\rowcolor{lightlightgray}~w/o Qry Neg   & 33.2               & 1.6          & 11.9           & 14.1                                      & 14.5 \\ \bottomrule

\end{tabular}
}

\vspace{-3.5mm}
\label{tab:qry-neg-ablation}
\end{table}

\custompara{Ablation on Contrastive Loss.}
Compared to standard contrastive loss, we introduce query images as hard negative examples during training.
To investigate the impact of this design, we train CoCa-based \ModelName-B without these hard negatives and report the results in Table~\ref{tab:qry-neg-ablation}.
As we can see, without query negatives, the performance of \ModelName drops across all benchmarks, significantly on the CIRR, CIRCO, and DTIN benchmarks.
Also, we find in many cases, this model prefers to rank the query image itself higher than other images during retrieval, regardless of the given instructions.
This indicates that differentiating closely similar images is crucial in improving the model's instruction understanding capabilities.
Importantly, although using query negatives seems to limit \ModelName' ability to find the identical image, the first example in Figure~\ref{fig:teaser-figure} shows \ModelName can generalize to this instruction unseen during training and successfully retrieve the identical image.

\custompara{Ablation on Model Architecture.}
\begin{table}[tb]
\small
\centering
\vspace{-2.5mm}
\caption{\label{tab:model-arch-comparison} Results of \ModelName variants. CrossAttn indicates the model with cross-attention instead of self-attention for modality integration. FrozenEnc means the model with backbone vision and language encoders frozen during training.}
\begin{tabular}{@{\;}l@{\;\;\;}c@{\;\;}c@{\;\;}c@{\;\;}c@{\;\;}c@{\;}}
\toprule
& \texttt{FIQ} & \texttt{CIRR} & \texttt{CIRCO} & \texttt{DTIN} & \texttt{GeneCIS} \\
& R@10               & R@1           & mAP@5          & R@10                             & R@1                         \\ \midrule
\rowcolor{lightgray}  \textbf{MagicLens-B} & \textbf{35.2} & \textbf{31.6} & \textbf{30.8} & \textbf{46.8} & \textbf{17.4} \\
w/ CrossAttn         & 31.0          & 28.3          & 27.0          & 41.4          & 16.2          \\
w/ FrozenEnc         & 30.8          & 25.9          & 21.7          & 30.1          & 15.2          \\ \midrule
\rowcolor{lightgray}  \textbf{MagicLens-L} & \textbf{38.0} & \textbf{33.3} & \textbf{34.1} & 48.2          & \textbf{16.7} \\
w/ CrossAttn         & 32.3          & 29.9          & 28.5          & \textbf{52.5} & 16.5          \\
w/ FrozenEnc         & 32.5          & 26.5          & 23.0          & 29.4          & 15.5          \\ \bottomrule
\end{tabular}

\vspace{-4.5mm}
\end{table}
We provide results of other model architectures we have explored in Table~\ref{tab:model-arch-comparison}.
In CrossAttn model arch, we explore various forms of cross attention, we report the best one which uses text embedding to attend concatenated image and text embeddings.
However, even the best variant of this arch fails to reach the performance of self attention on most benchmarks.

We also explore the impact of freezing the backbone encoders initialized from CoCa~\cite{Yu2022CoCa} during training.
The results of FrozenEnc are consistently worse than the fully-trained \ModelName.
This proves that merely training additional layers on the top of single-modality encoders is not sufficient to deliver the strongest model.

\subsection{Retrieval on 1.4M Open-Domain Image Corpus}
To simulate image retrieval in a more realistic scenario, we hold out 1.4M unseen images as our index set, making it the largest retrieval pool to date.
We then collect 150 images and divide them into three disjoint groups with different types of manually written instructions: simple, complex, and beyond visual.
Both simple and complex instructions are used in searching for visually similar images, but they differ in terms of complexity.
Simple instructions describe only one visual difference (e.g. \nlp{same product with different color}) in the images given, whereas complex ones have multiple differences (e.g., \nlp{car} and \nlp{bag} examples in Figure~\ref{fig:case-study-holdout}).
Beyond visual instructions aim to find images that share no visual similarities with the query images (e.g., \nlp{find other attractions}... in Figure~\ref{fig:teaser-figure}).

Table~\ref{tab:human-eval} compares CoCa-based \ModelName-L with code-available previous-best model (LinCIR;~\citet{Gu2023LinCIR}), both with ViT-L backbones.
For each query, one-on-one human evaluation is applied to the images retrieved by these two models to select the one that fully meets the instructions.
If both or neither of the models succeed, the evaluators will mark them as a tie.
We can observe LinCIR can handle simple instructions but suffers from complex instructions and almost completely fails on beyond visual instructions.
In contrast, our method can satisfy diverse search intents expressed by all kinds of instructions, remarkably on the complex (61.3 vs. 24.0) and beyond visual (80 vs. 4.7) ones.

\begin{table}[tb]
\centering
\small
\caption{One-on-one comparison (win rate) on a holdout index set with 1.4M images. Each setting has 50 queries with manually written instructions. The results are averaged over three evaluators. }
\begin{tabular}{c@{\quad}c@{\quad}c@{\quad}c}
\toprule
\textbf{Instruction Type} & \ModelName-L Win & LinCIR Win & Tie  \\ \midrule
\cellcolor[HTML]{fde0dd} Simple Visual & \textbf{50.7} & 41.3   & 8.0  \\
\cellcolor[HTML]{fbb4b9} Complex Visual & \textbf{61.3} & 24.0   & 14.7 \\
\cellcolor[HTML]{f768a1} Beyond Visual & \textbf{80.0} & 4.7    & 15.3 \\ \bottomrule
\end{tabular}
\vspace{-6mm}
\label{tab:human-eval}
\end{table}
\subsection{Qualitative Study}

\begin{figure}[tb]
    \centering
    % \vspace{-1mm}
    \includegraphics[width=0.96\linewidth]{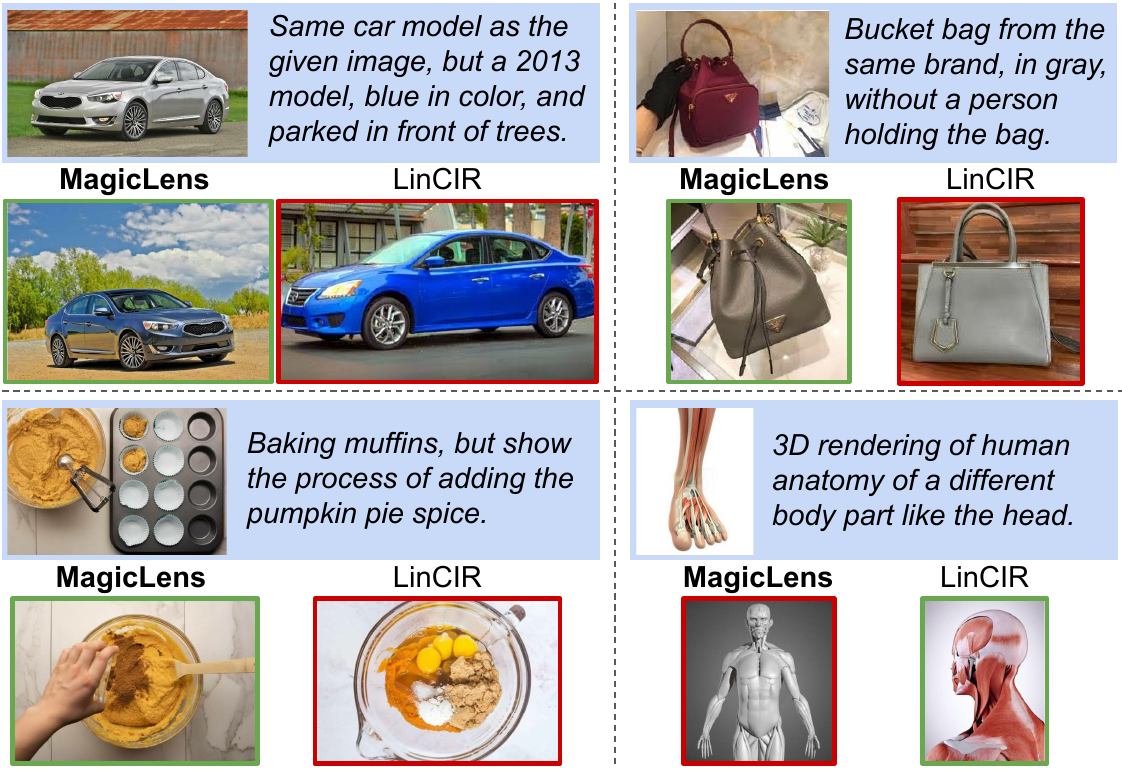}
    % \vspace{-1mm}
    \caption{Top-1 retrieved images of CoCa-based \ModelName-L and LinCIR on the holdout index set with 1.4M images.
    Queries are with a blue background, while correct and incorrect retrieved images are marked with green and red outlines, respectively.
    LinCIR fails to retrieve correct results for \texttt{car}, \texttt{bag}, and \texttt{muffin} queries, even considering its top-5 results (see Figure~\ref{appendix-fig:case-study-holdout-top5} in Appendix).
    }
    \vspace{-5mm}
    \label{fig:case-study-holdout}
\end{figure}

Figure~\ref{fig:case-study-holdout} illustrates top-1 retrieval results on the holdout index set with 1.4M images.
Even with complex instruction containing multiple conditions (\texttt{car} and \texttt{bag} examples),  \ModelName is still able to accurately comprehend search intents and retrieve desired images.
The \texttt{muffin} example showcases that \ModelName~can understand the non-trivial temporal relation between images, thanks to the relation diversity introduced by naturally occurring image pairs.
However, the image retrieved by \ModelName~given the \texttt{3D anatomy} query may not be generally preferred since the instruction exemplifies the head.
This suggests our model may return qualified yet imperfect examples when the instruction is not clearly expressed.
Please refer to Figure~\ref{appendix-fig:case-study-holdout-top5} in Appendix~\ref{appendix:more-qualitative-study} for more qualitative studies.

\begin{figure}[tbh]
    \centering
    \includegraphics[width=1.0\linewidth]{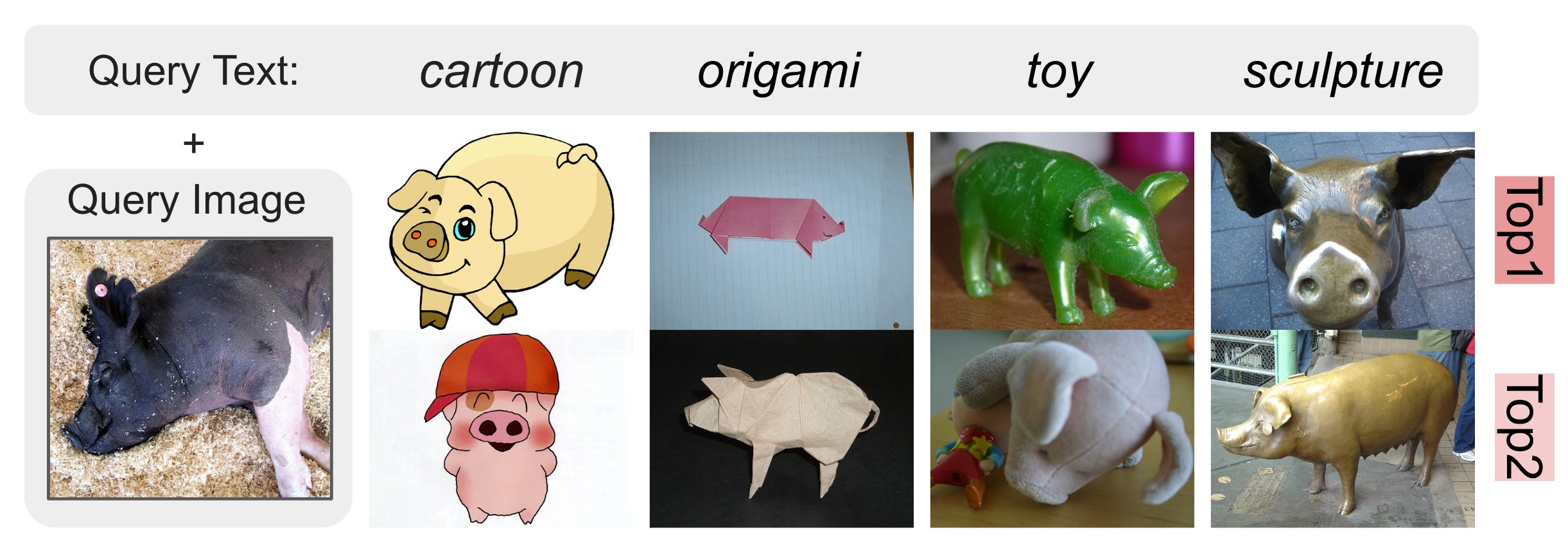}
    \vspace{-5.5mm}
    \caption{Top-2 retrieval results on four domains given the same query image on the DTIN benchmark.
    }
    \vspace{-5mm}
    \label{fig:case-study-domaintrans}
\end{figure}

Figure~\ref{fig:case-study-domaintrans} presents a visual case study on domain transfer retrieval using the DTIN benchmark.
The text instruction presented in each domain is \textit{``find this object in \{domain\}''}, where the same query image is used.
All top-2 retrieved results are correct, highlighting the effectiveness of \ModelName in understanding conceptual image relations.

\section{Conclusion}
We present \ModelName, a series of image retrieval models that follow open-ended text instructions.
Despite being 50$\times$ smaller than prior SOTA methods, \ModelName achieves better results on multiple benchmarks including CIRCO, GeneCIS, and DTIN.
Human evaluation on the 1.4M retrieval image pool shows that \ModelName can well satisfy diverse search intents expressed by open-ended instructions, especially complex and beyond visual ones.
This indicates \ModelName' strong capability and potential for real-world search scenarios.
Such retrieval models that support open-ended instructions can potentially benefit other vision-language tasks such as visual QA~\cite{Antol2015VQA, Chen2023InfoSeek}, and enhance multimodal retrieval augmented models~\cite{chen2022murag, hu2023reveal}.
More importantly, we hope our recipe for constructing large-scale synthetic self-supervised training data can shed light on other research directions, such as multimodal retrieval, multimodal representation learning, and beyond.

\section*{Impact Statement}
This work provides novel insights into self-supervised training by mining naturally occurring image pairs and develops image retrieval models that follow open-ended instructions to satisfy diverse search intents.
It may enable a wide range of search scenarios and have potentials for real-world applications by providing users with more accurate search results. 
However, despite careful filtering out explicit and offensive images in the training data, MagicLens' ability to understand image relationships could still be misused for inappropriate image searches.
Careful consideration and mitigation strategies are necessary to address these risks.

\section*{Acknowledgements}
The authors would like to thank Jinhyuk Lee, William Cohen, Jonathan Berant, Kristina Toutanova, Boqing Gong, and other members from the Google DeepMind Seattle for their constructive feedback.
% \pagebreak
% \newpage

\bibliography{reference}
\bibliographystyle{icml2024}

%%%%%%%%%%%%%%%%%%%%%%%%%%%%%%%%%%%%%%%%%%%%%%%%%%%%%%%%%%%%%%%%%%%%%%%%%%%%%%%
%%%%%%%%%%%%%%%%%%%%%%%%%%%%%%%%%%%%%%%%%%%%%%%%%%%%%%%%%%%%%%%%%%%%%%%%%%%%%%%
% APPENDIX
%%%%%%%%%%%%%%%%%%%%%%%%%%%%%%%%%%%%%%%%%%%%%%%%%%%%%%%%%%%%%%%%%%%%%%%%%%%%%%%
%%%%%%%%%%%%%%%%%%%%%%%%%%%%%%%%%%%%%%%%%%%%%%%%%%%%%%%%%%%%%%%%%%%%%%%%%%%%%%%
% \newpage
\appendix
\onecolumn
\section*{Overview of Appendix}
Our supplementary includes the following sections:
\begin{itemize}[leftmargin=*, nosep]
    \item \textbf{Section~\ref{appendix:implementation-details}: Implementation Details.}
    \item \textbf{Section~\ref{appendix:baselines}: Baselines.}
    \item \textbf{Section~\ref{appendix:detailed-comparison-results}: Full Results.}
    \item \textbf{Section~\ref{appendix:more-qualitative-study}: More Qualitative Study.}
\end{itemize}
\section{Implementation Details}
\label{appendix:implementation-details}
\custompara{Image Cleaning and Pairing.}
We use the Common Crawl and group images with identical URLs, considering them as images from the same websites.
We regard two images as identical if their CLIP image embedding scores exceed 0.98 and remove them.
If two groups share a high ratio of duplicated images (80\%), we randomly remove one of those groups.
The minimum resolution remained is 288x288, which matches the input size of CoCa models we used.
For the concrete thresholds used for filtering, we have set 0.82 as the threshold for CLIP image-to-image similarity and 0.9 for text-to-text similarity over captions.
Additionally, to ensure the uniqueness of the images, the target image must have a distinct ICA label with high text-image similarity to itself (0.32) and low similarity to the query image (0.18). 
Only image pairs that meet these requirements will be remained for the instruction generation stage.

\custompara{Instruction Generation.}
We provide LLMs with massive metadata expansion including Alt-texts, image content annotation (ICA) labels, and image captions by using various tools and LMMs.
Specifically, similar to~\citet{Sharma2018ConceptualCaptions}, we analyze candidate Alt-texts with part-of-speech, sentiment, and pornography annotations of Google Natural Language APIs.
We discard images if their Alt-texts only have rare tokens or if they are triggered by sentiment/pornography detectors
For ICA labels, we utilize Google Vision APIs to annotate entities for each image such as general objects, locations, and activities.
On average, each image has 25.2 fine-grained ICA labels.
Also, we provide the instruction and two detailed demonstrations for instruction generation in Table~\ref{appendix-tab:query-gen}.

\custompara{Model.} With the proposed data construction pipeline, we eventually collect $36,714,118$ triplets for pre-training.
For model architecture, we design 4 randomly initialized self-attention layers on the top of vision and language encoders.
Further, we utilize one attention pooling layer~\cite{Yu2022CoCa} for the final embedding.
Following~\citet{Jia2021ALIGN, Yu2022CoCa}, during the training of CoCa-based \ModelName, we set image resolution of 288$\times$288 and patch size 18$\times$18.
For CLIP-based \ModelName, we set image resolution of 224$\times$224 and use ViT-B16 and ViT-L14.
For both CLIP and CoCa, we use contrastive image embedding and contrastive text embedding, which will be concatenated as a sequence with a fixed length of 2 in self-attention layers.
The number of newly added self-attention layers is 4 and the $\tau$ is learnable and initialized with 0.07.
We set the batch size as 2048 and trained our models for a maximum of 50,000 steps with Adafactor~\cite{Shazeer2018Adafactor} and an early stopping mechanism.
The learning rates are set differently for newly-introduced parameters and re-used CLIP or CoCa parameters, at 2e-5 and 2e-6, respectively.
We train our base and large models on 64 and 128 TPUs, respectively. The training process lasts six hours for both models and the best checkpoints are selected based on the performance on the validation set of CIRR and CIRCO.

\section{Baselines}
\label{appendix:baselines}
\begin{table}[tb]

\centering
\small
\caption{Detailed prompt for query generation using PaLM2~\cite{Anil2023PaLM2}.}
\resizebox{0.98\linewidth}{!}{
\begin{tabular}{p{0.12\textwidth}|p{0.86\textwidth}}
\toprule
% \textbf{Type} & \textbf{Examples} \\ \midrule
Instruction  & \parbox{0.85\textwidth}{
Based on the provided ALT\_TEXT, TEXT\_LABEL, and CAPTION of two different images, create an interesting text query which can be used with the source image to retrieve the target image.\\
Note that the TEXT\_LABEL and CAPTION are generated by models so they may not be 100\% correct, especially when it's about very specific entities (e.g., a specific car type in some year), so selectively use the most likely correct information and generate the query.\\
This query should include:\\
1) one general and unspecific similarity (same brand, similar toy, similar building, etc).\\
2) all differences that only the target image has.\\
Remember the query should be concise, short, and not be able to directly retrieve the target image. The retrieval has to be done by combining the source image and text query.
}\\ \midrule
Demonstrations & \parbox{0.85\textwidth}{
Both images are from the website [ HOME - 1-of-1 Automotive Artworks ]\\
\textbf{Source Image}: ALT\_TEXT [custom porsche cayman gt4 illustration framed]. TEXT\_LABEL [Licence plate]. CAPTION [a drawing of a porsche gt4 rs coupe in a frame].\\
\textbf{Target Image}: ALT\_TEXT [custom illustration of a 1972 porsche 911 blue]. TEXT\_LABEL [Licence plate, Turquoise]. CAPTION [a framed print of a blue porsche 911 s all coupe].\\
\textbf{Think}: Both images are custom illustrations of Porsche cars as described in the alt\_text. The source image is a Porsche Cayman GT4 while the target image is a 1972 Porsche 911 in blue. Therefore, the query should focus on the type of image (custom illustration of a Porsche car), but specify the different model and year (1972 Porsche 911) and color (blue).\\
\textbf{Query}: [Porsche 911 in blue shown in the same illustrative way.]
\\
\\
Both Images are from the website [ Rapunzel Worksheet | Printable Worksheets and Activities for Teachers, Parents, Tutors and Homeschool Families ]\\
\textbf{Source Image}: ALT\_TEXT [tangled rapunzel color pages printable]. TEXT\_LABEL [Coloring book]. CAPTION [rapunzel in a boat with lanterns floating in the air coloring page].\\
\textbf{Target Image}: ALT\_TEXT [cool rapunzel and flynn flower hair coloring page]. TEXT\_LABEL [Coloring book, Floral design]. CAPTION [a black and white drawing of rapunzel tangled with long hair and flowers in her dress].\\
\textbf{Think}: Both images are coloring pages featuring Rapunzel as described in the alt\_text. The source image shows Rapunzel in a boat with lanterns, while the target image shows Rapunzel with Flynn and flowers in her hair. Therefore, the query should focus on the type of image (Rapunzel coloring page), but specify the different scene (Rapunzel with Flynn and flowers in her hair, not in a boat, no lanterns).\\
\textbf{Query}: [Same coloring page about Rapunzel but no boat or lantern, with more clear flowers in the character's hair]
\\
\\
\textit{…(Three more few-shot demonstrations)}

} \\ \bottomrule
\end{tabular}
% \vspace{-20em}
}
\label{appendix-tab:query-gen}

\end{table}
We consider various baselines and detail them as follows: \textbf{(1) PALARVA}~\cite{Cohen2022PALARVA}
\textbf{(2) Pic2Word}~\cite{Saito2023Pic2Word}, \textbf{(3) SEARLE}~\cite{Baldrati2023CIRCO}, \textbf{(4) ContextI2W}~\cite{Tang2023ContextI2W}, and \textbf{(5) LinCIR}~\cite{Gu2023LinCIR} train an additional mapping network to encode the given reference image as a pseudo word token.
Then, it can be combined with the actual query text for text-to-image retrieval.
These methods rely on image-caption pairs for mapping network training
Further, LinCIR introduces text-only data for better mapping capability.
\textbf{(6) CIReVL}~\cite{Karthik2023CIReVL} is a training-free method by using BLIP-2 with FLANT5-XXL~\cite{Li2023BLIP2} for query image caption generation, ChatGPT~\cite{OpenAI2022ChatGPT} for target image caption generation, and CLIP~\cite{Radford2021CLIP, Cherti2023OpenCLIP} for the final text-to-image retrieval.
Such a complex retrieval pipeline may limit their inference speed and potential practicalness in real-world scenarios.
Inspired by diffusion models, \textbf{(7) CompoDiff}~\cite{Gu2023CompoDiff} regards query text as a condition to guide the image embedding generation and train the model with 18M synthesized data.
\textbf{(8) PLI}~\cite{Chen2023PLI} corrupts the image in image-caption data and regards the original image as a target to simulate the CIR task during the pre-training stage.

\begin{table}[tbh]
\small
\centering
% \resizebox{0.96\linewidth}{!}
\caption{Detailed performance of CoCa-based \ModelName-B trained with IP2P data and our constructed data, on the same 1M scale.}
{
\begin{tabular}{l@{\;}c@{\;\;}c@{\;\;}c@{\;\;}c@{\;\;}c@{\;\;}c}
\toprule
         & \texttt{FIQ} & \texttt{CIRR} & \texttt{CIRR} & \texttt{CIRCO} & \texttt{DTIN} & \texttt{GeneCIS} \\
         & R@10               & R@1     & R$_s$@1      & mAP@5          & R@10                             & R@1                         \\ \midrule
% CLIP+IP2P$^*$ & 7.01	              & 4.07	      & 1.83		   & -                                & 2.44 \\ \midrule
\ModelName + IP2P (1M)     & 20.3               & 12.5     & 54.9      & 13.6            & 30.2                             & 14.5                                \\

\ModelName + Ours (1M)     & \textbf{33.5}               & \textbf{29.6}    & \textbf{66.9}      & \textbf{29.7}           & \textbf{43.7}                             & \textbf{15.8}                                 \\\bottomrule
\end{tabular}
}
\vspace{-1.5em}
\label{appendix-tab:ip2p-training-comparison}
\end{table}

\section{Full Results}
\label{appendix:detailed-comparison-results}

% \usepackage{multirow}
\begin{table*}[!b]
\small
\centering
\caption{Full results on the FIQ benchmark~\cite{Wu2021FashionIQ}.
CLIP-H and CLIP-G are OpenCLIP~\cite{Cherti2023OpenCLIP} checkpoints.
\textsuperscript{$\star$}CIReVL uses multiple model components, we omit ChatGPT and report \# parameters of other components (e.g., BLIP2-FLANT5-XXL + CLIP-G).
\textsuperscript{$\dag$}PLI does not release code so we estimate.
}
\resizebox{0.98\linewidth}{!}{
\begin{tabular}{lcccccccccc}
\toprule
\multicolumn{1}{c}{\multirow{2}{*}{\textbf{Method}}} & \multirow{2}{*}{\textbf{\begin{tabular}[c]{@{}c@{}} \# Total \\Params\end{tabular}}} & \multicolumn{1}{c}{\multirow{2}{*}{\textbf{\begin{tabular}[c]{@{}c@{}}Backbone\\ Network\end{tabular}}}} & \multicolumn{2}{c}{\textbf{Dress}}          & \multicolumn{2}{c}{\textbf{Shirt}}          & \multicolumn{2}{c}{\textbf{Toptee}}         & \multicolumn{2}{c}{\textbf{Overall}}        \\ \cmidrule(lr){4-5} \cmidrule(lr){6-7} \cmidrule(lr){8-9} \cmidrule(lr){10-11} 
\multicolumn{1}{c}{}                                 &                                    & \multicolumn{1}{c}{}                                                                                     & R@10                 & R@50                 & R@10                 & R@50                 & R@10                 & R@50                 & R@10                 & R@50                 \\ \midrule
PALAVRA \cite{Cohen2022PALARVA}                      & 176M\pz                            & CLIP-B                                                                                                   & 17.3                 & 35.9                 & 21.5                 & 37.1                 & 20.6                 & 38.8                 & 19.8                 & 37.3                 \\
SEARLE \cite{Baldrati2023CIRCO}                      & 165M\pz                            & CLIP-B                                                                                                   & 18.5                 & 39.5                 & 24.4                 & 41.6                 & 25.7                 & 46.5                 & 22.9                 & 42.5                 \\
CIReVL \cite{Karthik2023CIReVL} & 12.3B\textsuperscript{$\star$} & CLIP-B   & 25.3  & 46.4  & 28.4  & 47.8  & 31.2  & 53.9  & 28.3  & 49.4 \\
PLI  \cite{Chen2023PLI}                              & 224M\pz                            & BLIP-B                                                                                                   & 28.6                 & 50.8                 & 38.1                 & 57.8                 & 40.9                 & 62.7                 & \textbf{35.9}        & \textbf{57.1}        \\
\rowcolor{lightlightgray} \textbf{\ModelName-B}      & 166M\pz                            & CLIP-B                                                                                                   & 21.5                 & 41.3                 & 27.3                 & 48.8                 & 30.2                 & 52.3                 & 26.3                 & 47.4                 \\
\rowcolor{lightgray} \textbf{\ModelName-B}           & 267M\pz                            & CoCa-B                                                                                                   & 29.0                 & 48.9                 & 36.5                 & 55.5                 & 40.2                 & 61.9                 & 35.2                 & 55.4                 \\ \midrule
Pic2Word \cite{Saito2023Pic2Word}                    & 429M\pz                            & CLIP-L                                                                                                   & 20.0                 & 40.2                 & 26.2                 & 43.6                 & 27.9                 & 47.4                 & 24.7                 & 43.7                 \\
SEARLE \cite{Baldrati2023CIRCO}                      & 442M\pz                            & CLIP-L                                                                                                   & 20.5                 & 43.1                 & 26.9                 & 45.6                 & 29.3                 & 50.0                 & 25.6                 & 46.2                 \\
Context-I2W \cite{Tang2023ContextI2W}                & 496M\pz                            & CLIP-L                                                                                                   & 23.1                 & 45.3                 & 29.7                 & 48.6                 & 30.6                 & 52.9                 & 27.8                 & 48.9                 \\
CompoDiff \cite{Gu2023CompoDiff}                     & 568M\pz                            & CLIP-L                                                                                                   & 32.2                 & 46.3                 & 37.7                 & 49.1                 & 38.1                 & 50.6                 & 36.0        & 48.6                 \\
CIReVL \cite{Karthik2023CIReVL}  & 12.5B\textsuperscript{$\star$}  & CLIP-L  & 24.8  & 44.8  & 29.5  & 47.4  & 31.4  & 53.7  & 28.6  & 48.6 \\
PLI \cite{Chen2023PLI}                               & 428M\textsuperscript{$\dag$}       & CLIP-L                                                                                                   & 28.1                 & 51.1                 & 38.6                 & 58.5                 & 39.4                 & 62.7                 & 35.4                 & 57.4        \\
LinCIR \cite{Gu2023LinCIR}                           & 442M\pz                            & CLIP-L                                                                                                   & 20.9                 & 42.4                 & 29.1                 & 46.8                 & 28.8                 & 50.2                 & 26.3                 & 46.5                 \\
\rowcolor{lightlightgray} \textbf{\ModelName-L}      & 465M\pz                            & CLIP-L                                                                                                   & 25.5                 & 46.1                 & 32.7                 & 53.8                 & 34.0                 & 57.7                 & 30.7                 & 52.5        \\
\rowcolor{lightgray} \textbf{\ModelName-L}           & 613M\pz                            & CoCa-L                                                                                                   & 32.3                 & 52.7                 & 40.5                 & 59.2                 & 41.4                 & 63.0                 & \textbf{38.0}                 & \textbf{58.2}                 \\ \midrule
% TODO
Pic2Word \cite{Saito2023Pic2Word} & 987M\pz & CLIP-H & 28.0 & 51.5 & 36.9 & 56.0 & 40.2 & 62.0 & 35.0 & 56.5 \\
SEARLE \cite{Baldrati2023CIRCO}   & 1.0B\pz & CLIP-H & 28.5 & 51.1 & 36.5 & 55.5 & 38.8 & 60.9 & 34.6 & 55.8 \\
LinCIR \cite{Gu2023LinCIR}        & 1.0B\pz & CLIP-H & 29.8 & 52.1 & 36.9 & 57.8 & 42.1 & 62.5 & 36.3 & 57.5 \\
Pic2Word \cite{Saito2023Pic2Word} & 2.5B\pz & CLIP-G & 25.4 & 47.7 & 33.2 & 50.4 & 35.2 & 57.6 & 31.3 & 51.9 \\
SEARLE \cite{Baldrati2023CIRCO}   & 2.6B\pz & CLIP-G & 28.2 & 50.3 & 36.5 & 55.4 & 39.8 & 61.5 & 34.8 & 55.7 \\
CompoDiff \cite{Gu2023CompoDiff}  & 2.9B\pz & CLIP-G & 37.8 & 49.1 & 41.3 & 55.2 & 44.3 & 56.4 & 39.0 & 51.7 \\
CIReVL \cite{Karthik2023CIReVL}   & 14.6B\textsuperscript{$\star$} & CLIP-G  & 27.1  & 49.5  & 33.7  & 51.4  & 35.8  & 56.1  & 32.2  & 52.4 \\
LinCIR \cite{Gu2023LinCIR}        & 2.6B\pz & CLIP-G & 38.1 & 60.9 & 46.8 & 65.1 & 50.5 & 71.1 & \textbf{45.1} & \textbf{65.7} \\ \bottomrule

\end{tabular}
}
\label{appendix-tab:full-fiq}
\end{table*}

\begin{table*}[tbh]
\centering
\small
\caption{Full results on the CIRR benchmark~\cite{Liu2021CIRR}.
CLIP-H and CLIP-G are OpenCLIP~\cite{Cherti2023OpenCLIP} checkpoints.
\textsuperscript{$\star$}CIReVL uses multiple model components, we omit ChatGPT and report \# parameters of other components (e.g., BLIP2-FLANT5-XXL + CLIP-G).
\textsuperscript{$\dag$}PLI does not release code so we estimate.
}
\begin{tabular}{lccccccccc}
\toprule
\multicolumn{1}{c}{\multirow{2}{*}{\textbf{Method}}} & \multirow{2}{*}{\textbf{\begin{tabular}[c]{@{}c@{}} \# Total \\Params\end{tabular}}} & \multirow{2}{*}{\textbf{\begin{tabular}[c]{@{}c@{}}Backbone\\ Network\end{tabular}}} & \multicolumn{4}{c}{\textbf{Index Set}}                                                    & \multicolumn{3}{c}{\textbf{Subset}}                                \\ \cmidrule(lr){4-7} \cmidrule(lr){8-10}
\multicolumn{1}{c}{}                                 &                                     &                                                                                      & R@1                  & R@5                  & R@10                 & R@50                 & R@1                  & R@2                  & R@3                  \\ \midrule
PALAVRA \cite{Cohen2022PALARVA}                      & 176M\pz                             & CLIP-B                                                                               & 16.6                 & 43.5                 & 58.5                 & 84.0                 & 41.6                 & 65.3                 & 80.9                 \\
SEARLE \cite{Baldrati2023CIRCO}                      & 165M\pz                             & CLIP-B                                                                               & 24.0                 & 53.4                 & 66.8                 & 89.8                 & 54.9                 & 76.6                 & 88.2                 \\
CIReVL \cite{Karthik2023CIReVL}                      & 12.3B\textsuperscript{$\star$}            & CLIP-B                                                                               & 23.9                 & 52.5                 & 66.0                 & 87.0                 & 60.2                 & 80.1                 & 90.2                 \\
PLI \cite{Chen2023PLI}                               & 224M\textsuperscript{$\dag$}        & BLIP-B                                                                               & 27.2                 & 58.9                 & 71.4                 & 91.3                 & 55.1                 & 77.4                 & 89.1                 \\
\rowcolor{lightlightgray} \textbf{\ModelName-B}           & 166M\pz                             & CLIP-B                                                                                & 27.0	& 58.0	& 70.9	& 91.1	& 66.7	& 83.9	& 92.4 \\               
\rowcolor{lightgray} \textbf{\ModelName-B}           & 267M\pz                             & CoCa-B                                                                                & \textbf{31.6}        & \textbf{64.0}        & \textbf{76.9}        & \textbf{93.8}        & \textbf{69.3}                & \textbf{86.0}        & \textbf{94.0}                \\ \midrule
Pic2Word \cite{Saito2023Pic2Word}                    & 429M\pz                             & CLIP-L                                                                               & 23.9                 & 51.7                 & 65.3                 & 87.8                 & -                    & -                    & -                    \\
SEARLE \cite{Baldrati2023CIRCO}                      & 442M\pz                             & CLIP-L                                                                               & 24.2                 & 52.5                 & 66.3                 & 88.8                 & 53.8                 & 75.0                 & 88.2                 \\
Context-I2W \cite{Tang2023ContextI2W}                & 496M\pz                             & CLIP-L                                                                               & 25.6                 & 55.1                 & 68.5                 & 89.8                 & -                    & -                    & -                    \\
CompoDiff \cite{Gu2023CompoDiff}                     & 568M\pz                             & CLIP-L                                                                               & 18.2                 & 53.1                 & 70.8                 & 90.3                 & 57.4                 & 77.1                 & 87.9                 \\
CIReVL \cite{Karthik2023CIReVL}                      & 12.5B\textsuperscript{$\star$}            & CLIP-L                                                                               & 24.6                 & 52.3                 & 64.9                 & 86.3                 & 59.5                 & 79.9                 & 89.7                 \\
PLI \cite{Chen2023PLI}                               & 428M\textsuperscript{$\dag$}        & CLIP-L                                                                               & 25.5                 & 54.6                 & 67.6                 & 88.7                 & 55.6                 & 77.5                 & 89.5                 \\
LinCIR \cite{Gu2023LinCIR}                           & 442M\pz                             & CLIP-L                                                                               & 25.0                 & 53.3                 & 66.7                 & -                    & 57.1                 & 77.4                 & 88.9                 \\
\rowcolor{lightlightgray} \textbf{\ModelName-L}           & 465M\pz                             & CLIP-L  & 30.1	& 61.7	& 74.4	& 92.6	& 68.1	& 84.8	& 93.2 \\
\rowcolor{lightgray} \textbf{\ModelName-L}           & 613M\pz                             & CoCa-L                                                                                & \textbf{33.3}        & \textbf{67.0}        & \textbf{77.9}        & \textbf{94.4}        & \textbf{70.9}                & \textbf{87.3}        & \textbf{94.5} \\ \midrule
Pic2Word \cite{Saito2023Pic2Word}                    & 987M\pz                             & CLIP-H                                                                               & 32.9                 & 63.1                 & 73.9                 & -                    & 62.2                 & 81.4                 & 91.2                 \\
SEARLE \cite{Baldrati2023CIRCO}                      & 1.0B\pz                             & CLIP-H                                                                               & 34.0                 & 64.0                 & 75.3                 & -                    & 64.6                 & 83.2                 & 92.8                 \\
LinCIR \cite{Gu2023LinCIR}                           & 1.0B\pz                             & CLIP-H                                                                               & 33.8                 & 63.5                 & 73.4                 & -                    & 62.4                 & 81.5                 & 92.1                 \\
Pic2Word \cite{Saito2023Pic2Word}                    & 2.5B\pz                             & CLIP-G                                                                               & 30.4                 & 58.1                 & 69.2                 & -                    & 68.9        & 85.5        & 93.0                 \\
SEARLE \cite{Baldrati2023CIRCO}                      & 2.6B\pz                             & CLIP-G                                                                               & 34.8        & 64.1                 & 75.1                 & -                    & 68.7                 & 84.7                 & 93.2        \\
CompoDiff \cite{Gu2023CompoDiff}                     & 2.9B\pz                             & CLIP-G                                                                               & 26.7                 & 55.1                 & 74.5                 & 92.0        & 64.5                 & 82.4                 & 91.8                 \\
CIReVL \cite{Karthik2023CIReVL}                      & 14.6B\textsuperscript{$\star$}            & CLIP-G                                                                               & 34.7                 & 64.3                 & 75.1                 & 91.7                 & 68.0                 & 84.9                 & 93.2        \\
LinCIR \cite{Gu2023LinCIR}                           & 2.6B\pz                             & CLIP-G                                                                               & \textbf{35.3}                 & 64.7        & 76.1        & -                    & 63.4                 & 82.2                 & 92.0                \\ \bottomrule
\end{tabular}

\label{appendix-tab:full-cirr}
\end{table*}

\begin{table*}[tbh]
\small
\centering
\caption{Full results on the CIRCO benchmark~\cite{Baldrati2023CIRCO}.
CLIP-H and CLIP-G are OpenCLIP~\cite{Cherti2023OpenCLIP} checkpoints.
\textsuperscript{$\star$}CIReVL uses multiple model components, we omit ChatGPT and report \# parameters of other components (e.g., BLIP2-FLANT5-XXL + CLIP-G).
\textsuperscript{$\dag$}PLI does not release code so we estimate.
}
\begin{tabular}{lcccccc}
\toprule
\multicolumn{1}{c}{\textbf{Method}}        & \textbf{\begin{tabular}[c]{@{}c@{}} \# Total \\Params\end{tabular}}           & \textbf{\begin{tabular}[c]{@{}c@{}}Backbone\\ Network\end{tabular}} & mAP@5                & mAP@10               & mAP@25               & mAP@50               \\ \midrule
PALAVRA \cite{Cohen2022PALARVA}            & 176M\pz                      & CLIP-B                                                              & 4.6                  & 5.3                  & 6.3                  & 6.8                  \\
SEARLE \cite{Baldrati2023CIRCO}            & 165M\pz                      & CLIP-B                                                              & 9.4                  & 9.9                  & 11.1                 & 11.8                 \\
CIReVL \cite{Karthik2023CIReVL}            & 12.3B\textsuperscript{$\star$}     & CLIP-B                                                              & 14.9                 & 15.4                 & 17.0                 & 17.8                 \\
PLI \cite{Chen2023PLI}                     & 224M\textsuperscript{$\dag$} & BLIP-B                                                              & 7.1                  & 8.0                  & 9.2                  & 9.7                  \\
\rowcolor{lightlightgray} \textbf{\ModelName-B} & 166M\pz                      & CLIP-B                                                              & 23.1	& 23.8	& 25.8	& 26.7\\
\rowcolor{lightgray} \textbf{\ModelName-B} & 267M\pz                      & CoCa-B                                                              & \textbf{30.8}        & \textbf{32.0}        & \textbf{34.5}        & \textbf{35.6}        \\ \midrule
Pic2Word \cite{Saito2023Pic2Word}          & 429M\pz                      & CLIP-L                                                              & 8.7                  & 9.5                  & 10.6                 & 11.3                 \\
SEARLE \cite{Baldrati2023CIRCO}            & 442M\pz                      & CLIP-L                                                              & 11.7                 & 12.7                 & 14.3                 & 15.1                 \\
CompoDiff \cite{Gu2023CompoDiff}           & 568M\pz                      & CLIP-L                                                              & 12.6                 & 13.4                 & 15.8                 & 16.4                 \\
CIReVL \cite{Karthik2023CIReVL}            & 12.5B\textsuperscript{$\star$}     & CLIP-L                                                              & 18.6                 & 19.0                 & 20.9                 & 21.8                 \\
PLI \cite{Chen2023PLI}                     & 428M\textsuperscript{$\dag$} & CLIP-L                                                              & 10.4                 & 11.6                 & 13.0                 & 13.7                 \\
LinCIR \cite{Gu2023LinCIR}                 & 442M\pz                      & CLIP-L                                                              & 12.6                 & 13.6                 & 15.0                 & 15.9                 \\
% 29.6	30.8	33.4	34.4
\rowcolor{lightlightgray} \textbf{\ModelName-L} & 465M\pz                      & CLIP-L                                                              & 29.6 & 30.8 & 33.4 & 34.4 \\ 
\rowcolor{lightgray} \textbf{\ModelName-L} & 613M\pz                      & CoCa-L                                                              & \textbf{34.1}        & \textbf{35.4}        & \textbf{38.1}        & \textbf{39.2}        \\ \midrule
Pic2Word \cite{Saito2023Pic2Word}          & 987M\pz                      & CLIP-H                                                              & 11.7                 & 12.3                 & 13.7                 & 14.4                 \\
SEARLE \cite{Baldrati2023CIRCO}            & 1.0B\pz                      & CLIP-H                                                              & 16.1                 & 16.9                 & 18.8                 & 19.7                 \\
LinCIR \cite{Gu2023LinCIR}                 & 1.0B\pz                      & CLIP-H                                                              & 17.6                 & 18.5                 & 20.5                 & 21.4                 \\
Pic2Word \cite{Saito2023Pic2Word}          & 2.5B\pz                      & CLIP-G                                                              & 5.5                  & 5.6                  & 6.7                  & 7.1                  \\
SEARLE \cite{Baldrati2023CIRCO}            & 2.6B\pz                      & CLIP-G                                                              & 13.2                 & 13.9                 & 15.3                 & 16.0                 \\
CompoDiff \cite{Gu2023CompoDiff}           & 2.9B\pz                      & CLIP-G                                                              & 15.3                 & 17.7                 & 19.4                 & -                    \\
CIReVL \cite{Karthik2023CIReVL}            & 14.6B\textsuperscript{$\star$}     & CLIP-G                                                              & 26.8                 & 27.6                 & 30.0                 & 31.0                 \\
LinCIR \cite{Gu2023LinCIR}                 & 2.6B\pz                      & CLIP-G                                                              & 19.7                 & 21.0                 & 23.1                 & 24.2                 \\ \bottomrule
\end{tabular}
\label{appendix-tab:full-circo}
\end{table*}
\begin{table*}[tbh]
\caption{Full results on the DTIN benchmark~\cite{Saito2023Pic2Word}.
CLIP-G is a OpenCLIP~\cite{Cherti2023OpenCLIP} checkpoint.
\textsuperscript{$\star$}CIReVL uses multiple model components, we omit ChatGPT and report \# parameters of other components (e.g., BLIP2-FLANT5-XXL + CLIP-G).
}
\resizebox{0.98\linewidth}{!}{
\begin{tabular}{lcccccccccccc}
\toprule
\multicolumn{1}{c}{\multirow{2}{*}{\textbf{Method}}} & \multirow{2}{*}{\textbf{\begin{tabular}[c]{@{}c@{}} \# Total \\Params\end{tabular}}} & \multirow{2}{*}{\textbf{\begin{tabular}[c]{@{}c@{}}Backbone\\ Network\end{tabular}}} & \multicolumn{2}{c}{\textbf{Cartoon}} & \multicolumn{2}{c}{\textbf{Origami}} & \multicolumn{2}{c}{\textbf{Toy}} & \multicolumn{2}{c}{\textbf{Sculpture}} & \multicolumn{2}{c}{\textbf{Overall}} \\ \cmidrule(lr){4-5} \cmidrule(lr){6-7} \cmidrule(lr){8-9} \cmidrule(lr){10-11} \cmidrule(lr){12-13}
\multicolumn{1}{c}{}                                 &                                    &                                                                                      & R@10              & R@50             & R@10              & R@50             & R@10            & R@50           & R@10               & R@50              & R@10              & R@50             \\ \midrule
Image-only \cite{Saito2023Pic2Word}                                          & 304M\pz                            & CLIP-L                                                                               & 0.3               & 4.5              & 0.2               & 1.8              & 0.6             & 5.7            & 0.3                & 4.0               & 0.4               & 4.0              \\
Text-only \cite{Saito2023Pic2Word}                                           & 124M\pz                            & CLIP-L                                                                               & 0.2               & 1.1              & 0.8               & 3.7              & 0.8             & 2.4            & 0.4                & 2.0               & 0.5               & 2.3              \\
Image+Text \cite{Saito2023Pic2Word}                                          & 428M\pz                            & CLIP-L                                                                               & 2.2               & 13.3             & 2.0               & 10.3             & 1.2             & 9.7            & 1.6                & 11.6              & 1.7               & 11.2             \\ \midrule
Pic2Word \cite{Saito2023Pic2Word}                    & 429M\pz                            & CLIP-L                                                                               & 8.0               & 21.9             & 13.5              & 25.6             & 8.7             & 21.6           & 10.0               & 23.8              & 10.1              & 23.2             \\
Context-I2W \cite{Tang2023ContextI2W}                & 496M\pz                            & CLIP-L                                                                               & 10.2              & 26.1             & 17.5              & 28.7             & 11.6            & 27.4           & 12.1               & 28.2              & 12.9              & 27.6             \\
CIReVL \cite{Karthik2023CIReVL}                      & 14.6B\textsuperscript{$\star$}           & CLIP-G                                                                               & 19.2              & 42.8             & 30.2              & 41.3             & 22.2            & 43.1           & 23.4               & 45.0              & 23.8              & 43.0             \\ \midrule
\rowcolor{lightgray} \textbf{\ModelName-B}           & 166M\pz                           & CLIP-B                                                                                                        & 49.4	& 67.0	& 13.8	& 26.3	& 25.8	& 43.4	& 24.3	& 41.3	& 28.3	& 44.5           \\
\rowcolor{lightgray} \textbf{\ModelName-B}           & 267M\pz                           & CoCa-B                                                                                                        & 65.8              & 73.3             & 29.3              & 38.6             & 46.7            & 57.7           & 45.3               & 57.1              & 46.8              & 56.7            \\
\rowcolor{lightgray} \textbf{\ModelName-L}           & 465M\pz  & CLIP-L & 62.6 & 72.2 & 21.5	& 33.4	& 43.8 & 58.4 & 38.0 & 54.2 & 41.5 & 54.5 \\    
\rowcolor{lightgray} \textbf{\ModelName-L}           & 613M\pz                            & CoCa-L                                                                               & 60.1              & 69.6             & 36.0              & 44.7             & 45.2            & 56.9           & 51.4               & 59.6              & \textbf{48.2}     & \textbf{57.7}    \\ \bottomrule
\end{tabular}
}
\label{appendix-tab:full-dt}
\end{table*}

\begin{table*}[tbh]
\centering
\caption{Full results on the GeneCIS benchmark~\cite{Vaze2023GeneCIS}.
\textsuperscript{$\star$}CIReVL uses multiple model components, we omit ChatGPT and report \# parameters of other components (e.g., BLIP2-FLANT5-XXL + CLIP-G).
}
\resizebox{0.98\linewidth}{!}{
\begin{tabular}{lccccccccccccccc}
\toprule
\multicolumn{1}{c}{\multirow{2}{*}{\textbf{Method}}} & \multirow{2}{*}{\textbf{\# Params}} & \multirow{2}{*}{\textbf{\begin{tabular}[c]{@{}c@{}}Backbone\\ Network\end{tabular}}} & \multicolumn{3}{c}{\textbf{Focus Attribute}}                       & \multicolumn{3}{c}{\textbf{Change Attribute}}                      & \multicolumn{3}{c}{\textbf{Focus Object}}                          & \multicolumn{3}{c}{\textbf{Change Object}}                         & \textbf{Avg}         \\  \cmidrule(lr){4-6} \cmidrule(lr){7-9} \cmidrule(lr){10-12} \cmidrule(lr){13-15} \cmidrule(lr){16-16}
\multicolumn{1}{c}{}                                 &                                     &                                                                                      & R@1                  & R@2                  & R@3                  & R@1                  & R@2                  & R@3                  & R@1                  & R@2                  & R@3                  & R@1                  & R@2                  & R@3                  & R@1                  \\ \midrule
CIReVL (\citeyear{Karthik2023CIReVL})  & 12.3B\textsuperscript{$\star$}  & CLIP-B  & 17.9 & 29.4 & 40.4 & 14.8 & 25.8 & 35.8 & 14.6 & 24.3 & 33.3 & 16.1 & 27.8 & 37.6 & 15.9 \\
\rowcolor{lightgray} \textbf{\ModelName-B}           & 166M\pz                             & CLIP-B                                                                               & 15.5	& 28.4	& 39.1	& 12.3	& 23.0	& 32.1	& 14.4	& 26.2	& 35.5	& 17.7	& 28.4	& 39.2	& 15.0\\ 
\rowcolor{lightgray} \textbf{\ModelName-B}           & 267M\pz                             & CoCa-B                                                                               & 16.2                 & 27.8                 & 38.6                 & 16.2                 & 27.2                 & 36.6                 & 17.1                 & 27.7                 & 38.2                 & 20.2                 & 32.2                 & 42.9                 & \textbf{17.4}               \\ \midrule
Pic2Word (\citeyear{Saito2023Pic2Word})                & 429M\pz                             & CLIP-L                                                                               & 15.7                 & 28.2                 & 38.7                 & 13.9                 & 24.7                 & 33.1                 & 8.4                  & 18.0                 & 25.8                 & 6.7                  & 15.1                 & 24.0                 & 11.2                 \\
SEARLE (\citeyear{Baldrati2023CIRCO})                  & 442M\pz                             & CLIP-L                                                                               & 17.0                 & 29.7                 & 40.7                 & 16.4                 & 25.3                 & 34.1                 & 8.0                  & 16.9                 & 25.6                 & 7.9                  & 16.8                 & 24.8                 & 12.3                 \\
CompoDiff (\citeyear{Gu2023CompoDiff})                 & 568M\pz                             & CLIP-L                                                                               & 13.5                 & 24.3                 & 36.1                 & 19.2                 & 28.6                 & 37.2                 & 8.1                  & 16.4                 & 25.1                 & 18.7                 & 31.7                 & 40.6                 & 14.9                 \\
CIReVL (\citeyear{Karthik2023CIReVL})  & 12.5B\textsuperscript{$\star$}  & CLIP-L  & 19.5 & 31.8 & 42.0 & 14.4 & 26.0 & 35.2 & 12.3 & 21.8 & 30.5 & 17.2 & 28.9 & 37.6 & 15.9 \\
LinCIR (\citeyear{Gu2023LinCIR})                       & 442M\pz                             & CLIP-L                                                                               & 16.9                 & 30.0                 & 41.5                 & 16.2                 & 28.0                 & 36.8                 & 8.3                  & 17.4                 & 26.2                 & 7.4                  & 15.7                 & 25.0                 & 12.2                 \\ 
\rowcolor{lightgray} \textbf{\ModelName-L}           & 465M\pz                             & CLIP-L                                                                               & 16.1 & 28.2	& 39.0	& 15.6	& 27.5	& 36.3	& 16.3	& 26.2	& 35.5	& 17.1	& 29.5	& 39.7	& 16.3\\
\rowcolor{lightgray} \textbf{\ModelName-L}           & 613M\pz                             & CoCa-L                                                                               & 16.6                 & 28.7                 & 39.3                 & 16.0                 & 27.5                 & 36.5                 & 15.7                 & 27.6                 & 37.3                 & 18.7                 & 31.7                 & 40.2                 & \textbf{16.7}        \\ \midrule
Pic2Word (\citeyear{Saito2023Pic2Word})                & 987M\pz                             & CLIP-H                                                                               & 18.6                 & 30.7                 & 42.1                 & 13.2                 & 23.9                 & 33.1                 & 9.2                  & 17.6                 & 27.1                 & 6.6                  & 16.5                 & 25.4                 & 11.9                 \\
SEARLE (\citeyear{Baldrati2023CIRCO})                  & 1.0B\pz                             & CLIP-H                                                                               & 18.8                 & 31.5                 & 42.3                 & 15.5                 & 26.9                 & 35.9                 & 10.6                 & 18.7                 & 26.5                 & 8.5                  & 17.9                 & 26.2                 & 13.3                 \\
LinCIR (\citeyear{Gu2023LinCIR})                      & 1.0B\pz                             & CLIP-H                                                                               & 19.6                 & 31.5                 & 41.6                 & 16.6                 & 27.6                 & 37.5                 & 9.8                  & 18.8                 & 27.9                 & 9.0                  & 17.6                 & 25.7                 & 13.8                 \\
Pic2Word (\citeyear{Saito2023Pic2Word})                & 2.5B\pz                             & CLIP-G                                                                               & 12.5                 & 23.4                 & 33.7                 & 11.7                 & 21.9                 & 30.9                 & 9.9                  & 19.3                 & 27.4                 & 8.6                  & 18.2                 & 26.1                 & 10.7                 \\
SEARLE (\citeyear{Baldrati2023CIRCO})                  & 2.6B\pz                             & CLIP-G                                                                               & 16.3                 & 29.4                 & 40.7                 & 16.2                 & 27.3                 & 35.5                 & 10.8                 & 18.2                 & 27.9                 & 8.3                  & 15.6                 & 25.8                 & 12.9                 \\
CompoDiff (\citeyear{Gu2023CompoDiff})                 & 2.9B\pz                             & CLIP-G                                                                               & 14.3                 & 26.7                 & 38.4                 & 19.7                 & 28.8                 & 37.4                 & 9.2                  & 19.1                 & 25.8                 & 18.7                 & 31.7                 & 40.2                 & 15.5                 \\
% CIReVL (\citeyear{Karthik2023CIReVL})                  & 14.6B\textsuperscript{$\star$}            & CLIP-G                                                                               & 17.9                 & 29.4                 & 40.4                 & 14.8                 & 25.8                 & 35.8                 & 14.6                 & 24.3                 & 33.3                 & 16.1                 & 27.8                 & 37.6                 & 15.9                 \\
CIReVL (\citeyear{Karthik2023CIReVL})                  & 14.6B\textsuperscript{$\star$}            & CLIP-G                                                                               & 20.5                 & 34.0                 & 44.5                 & 16.1                 & 28.6                 & 39.4                 & 14.7                 & 25.2                 & 33.0                 & 18.1                 & 31.2                 & 41.0                 & \textbf{17.4}                 \\
LinCIR (\citeyear{Gu2023LinCIR})                       & 2.6B\pz                             & CLIP-G                                                                               & 19.1                 & 33.0                 & 42.3                 & 17.6                 & 30.2                 & 38.1                 & 10.1                 & 19.1                 & 28.1                 & 7.9                  & 16.3                 & 25.7                 & 13.7                 \\ \bottomrule
\end{tabular}
}
\label{appendix-tab:full-cond-img-sim}
\end{table*}

\subsection{Results on Five Multimodality-to-Image Benchmarks}
Table~\ref{appendix-tab:full-fiq}, \ref{appendix-tab:full-cirr}, and \ref{appendix-tab:full-circo} show the full results on three CIR benchmarks~\cite{Wu2021FashionIQ, Liu2021CIRR, Baldrati2023CIRCO}.
We report the performances of various models on \texttt{DT} and \texttt{GeneCIS} on Table~\ref{appendix-tab:full-dt} and Table~\ref{appendix-tab:full-cond-img-sim}, respectively.
Some prior methods may use larger encoders~\cite{Gu2023CompoDiff, Gu2023LinCIR} and develop a retrieval pipeline~\cite{Karthik2023CIReVL} including LLMs~\cite{OpenAI2022ChatGPT} and LMMs~\cite{Li2023BLIP2} for performance gains.
Despite this, their results are still worse than that of \ModelName, supporting the parameter efficiency claimed in Figure~\ref{fig:params-efficiency}.

\subsection{Data Training Comparison}

Table~\ref{appendix-tab:ip2p-training-comparison} compares CoCa-based \ModelName-B trained on IP2P data and on our data in detail, both at a 1M scale.
\begin{figure}[!h]
    \centering
    % \vspace{-1mm}
    \includegraphics[width=0.7\linewidth]{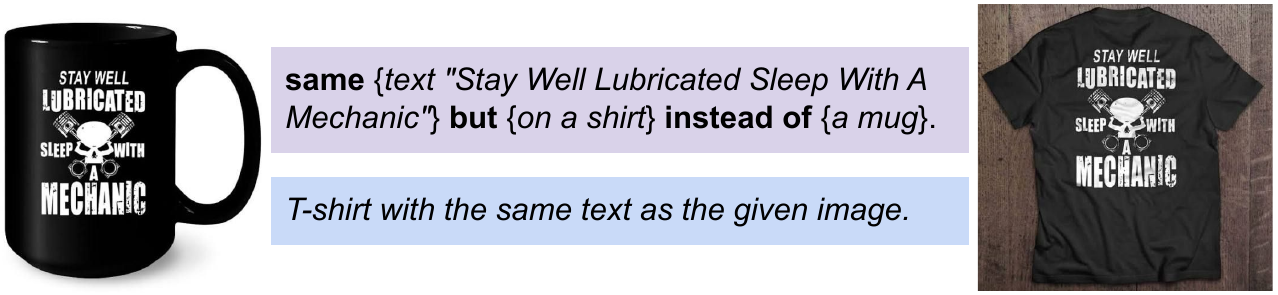}
    % \vspace{-1mm}
    \caption{
    Examples of \hlPurple{template-based} and \hlBlue{template-free} instructions for the same image pair.
    }
    % \vspace{-3mm}
    \label{appendix-fig:instruction-comparison}
\end{figure}

\subsection{Text-to-Image Retrieval with CLIP-based MagicLens}
\begin{table*}[t]
\centering
\small
\vspace{-1em}
\caption{\label{appendix-tab:clip-zs-it-rt} Zero-shot image-text retrieval results. Results are marked in bold if they are better than initialized checkpoints. \textsuperscript{$\star$}CLIP and CoCa we reproduced and used for \ModelName. \textsuperscript{$\dagger$}CLIP and CoCa reported in the original paper.}
% \resizebox{1.0\textwidth}{!}{%
% \begin{tabular}{@{}lcccccccccccc@{}}
\begin{tabular}{lcccccccccccc}
    \toprule 
    \multicolumn{1}{c}{\multirow{3}{*}{\textbf{Model}}} & \multicolumn{6}{c}{\texttt{Flickr30K} (1K test set)} & \multicolumn{6}{c}{\texttt{MSCOCO} (5K test set)} \\
    & \multicolumn{3}{c}{Image $\rightarrow$ Text} & \multicolumn{3}{c}{Text $\rightarrow$ Image} & \multicolumn{3}{c}{Image $\rightarrow$ Text} & \multicolumn{3}{c}{Text $\rightarrow$ Image} \\
    \cmidrule(lr){2-4} \cmidrule(lr){5-7} \cmidrule(lr){8-10} \cmidrule(lr){11-13}
     & R@1 & R@5 & R@10 & R@1 & R@5 & R@10 & R@1 & R@5 & R@10 & R@1 & R@5 & R@10 \\
    \midrule
    CLIP-B\textsuperscript{$\star$} & 81.7 & 97.1 & 98.5 & 61.6 & 85.6 & 91.2 & 51.9 & 76.3 & 83.9 & 32.1 & 56.7 & 67.6 \\
    \rowcolor{lightgray} \textbf{\ModelName-B} & 78.9 & 94.9 & 97.5 & \textbf{67.8} & \textbf{88.8} & \textbf{93.4} & 49.5 & 74.5 & 82.5 & \textbf{40.1} & \textbf{65.4} & \textbf{75.1} \\ \midrule

        % CoCa-Base & 89.8 & 98.8 & 99.8 & 76.8 & 93.7 & 96.8 & 63.8 & 84.7 & 90.7 & 47.5 & 72.4 & 80.9 \\
    \textcolor{gray}{CoCa-B\textsuperscript{$\dagger$}} & \textcolor{gray}{89.8} & \textcolor{gray}{98.8} & \textcolor{gray}{99.8} & \textcolor{gray}{76.8} & \textcolor{gray}{93.7} & \textcolor{gray}{96.8} & \textcolor{gray}{63.8} & \textcolor{gray}{84.7} & \textcolor{gray}{90.7} & \textcolor{gray}{47.5} & \textcolor{gray}{72.4} & \textcolor{gray}{80.9} \\
    CoCa-B\textsuperscript{$\star$} & 88.6 & 98.5 & 99.4 & 74.5 & 93.4 & 96.4 & 63.4 & 84.2 & 90.4 & 46.4 & 71.5 & 80.1 \\
    \rowcolor{lightgray} \textbf{\ModelName-B} & 87.9 & 97.7 & \textbf{99.5} & \textbf{76.2} & \textbf{93.7} & \textbf{96.5} & \textbf{64.8} & \textbf{85.5} & \textbf{91.2} & \textbf{48.9} & \textbf{73.9} & \textbf{82.5} \\ \midrule
    \textcolor{gray}{CLIP-L\textsuperscript{$\dagger$}} & \textcolor{gray}{88.0} & \textcolor{gray}{98.7} & \textcolor{gray}{99.4} & \textcolor{gray}{68.7} & \textcolor{gray}{90.6} & \textcolor{gray}{95.2} & \textcolor{gray}{58.4} & \textcolor{gray}{81.5} & \textcolor{gray}{88.1} & \textcolor{gray}{37.8} & \textcolor{gray}{64.2} & \textcolor{gray}{72.2} \\
    % 88	98.7	99.4	68.7	90.6	95.2	58.4	81.5	88.1	37.8	62.4	72.2
    CLIP-L\textsuperscript{$\star$} & 84.6 & 97.9 & 99.3 & 65.4 & 87.6 & 92.9 & 56.2 & 79.3 & 87.3 & 34.6 & 59.4 & 69.8\\
    \rowcolor{lightgray} \textbf{\ModelName-L} & 84.6 & 96.2 & 98.8 & \textbf{72.5} & \textbf{91.5} & \textbf{95.2} & 55.9 & 78.7 & 86.3 & \textbf{44.3} & \textbf{69.4} & \textbf{78.3} \\
    \midrule
    % CoCa-Large & 91.4 & 99.2 & 99.9 & 79.0 & 95.1 & 97.4 & 65.4 & 85.6 & 91.4 & 50.1 & 73.8 & 81.8 \\
    \textcolor{gray}{CoCa-L\textsuperscript{$\dagger$}} & \textcolor{gray}{91.4} & \textcolor{gray}{99.2} & \textcolor{gray}{99.9} & \textcolor{gray}{79.0} & \textcolor{gray}{95.1} & \textcolor{gray}{97.4} & \textcolor{gray}{65.4} & \textcolor{gray}{85.6} & \textcolor{gray}{91.4} & \textcolor{gray}{50.1} & \textcolor{gray}{73.8} & \textcolor{gray}{81.8} \\
    CoCa-L\textsuperscript{$\star$} & 92.1 & 98.8 & 99.9 & 78.4 & 94.2 & 96.9 & 65.1 & 85.5 & 91.3 & 49.3 & 73.2 & 81.5\\
    \rowcolor{lightgray} \textbf{\ModelName-L} & 89.6 & 98.7 & 99.4 & \textbf{79.7} & \textbf{95.0} & \textbf{97.4} & \textbf{67.7} & \textbf{87.6} & \textbf{92.7} & \textbf{53.1} & \textbf{77.4} & \textbf{84.9}\\ \bottomrule

\end{tabular}
% }

\vspace{-1.5em}
\end{table*}
We list the text-to-image retrieval results in Table~\ref{appendix-tab:clip-zs-it-rt} with the original CLIP and updated backbone CLIP encoders of \ModelName.
The text-to-image retrieval performance is significantly boosted on both base and large models where image-to-text retrieval ability marginally drops.
This aligns with the conclusion we draw on CoCa in \S~\ref{sec:text-to-image-retrieval}.

\subsection{Examples of Template-based Instructions}

We provide a concrete example of different instructions on the same image pair in Figure~\ref{appendix-fig:instruction-comparison}.

% \pagebreak
% \newpage
\section{More Qualitative Study}
\label{appendix:more-qualitative-study}
\begin{figure}[tbh]
    \centering
    \includegraphics[width=0.91\linewidth]{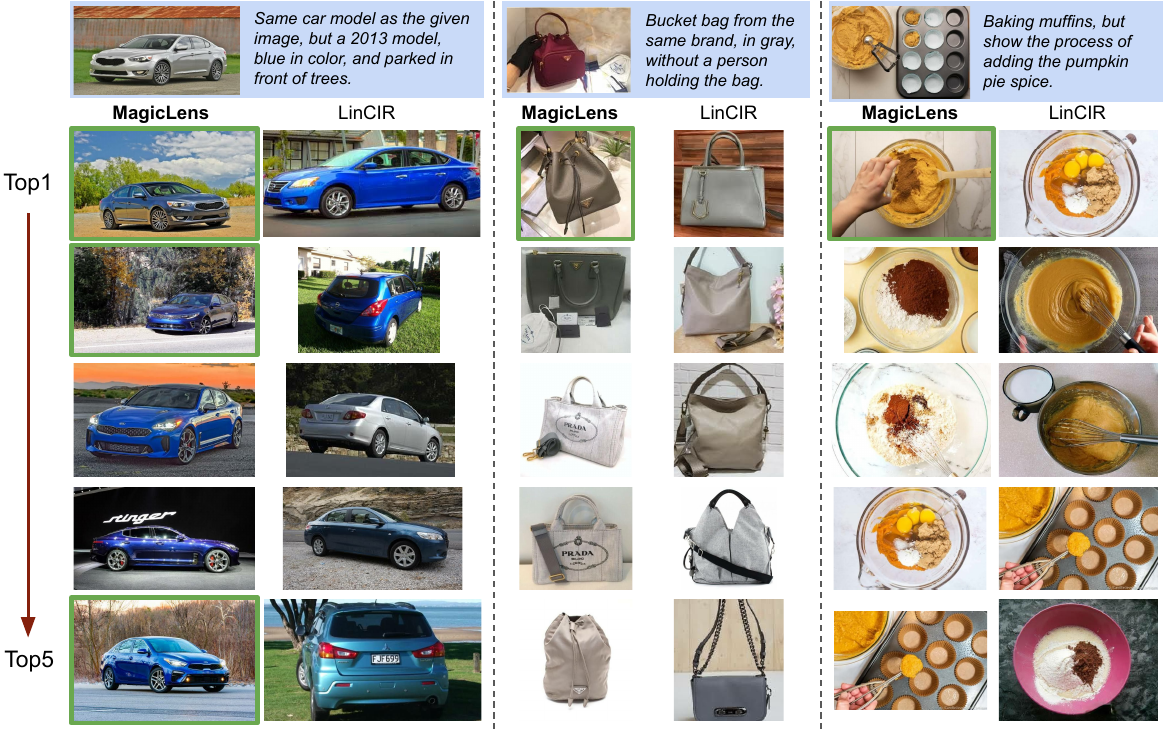}
    \vspace{1pt}
    \rule{0.92\linewidth}{0.5pt}
    \vspace{1pt}
    \includegraphics[width=0.91\linewidth]{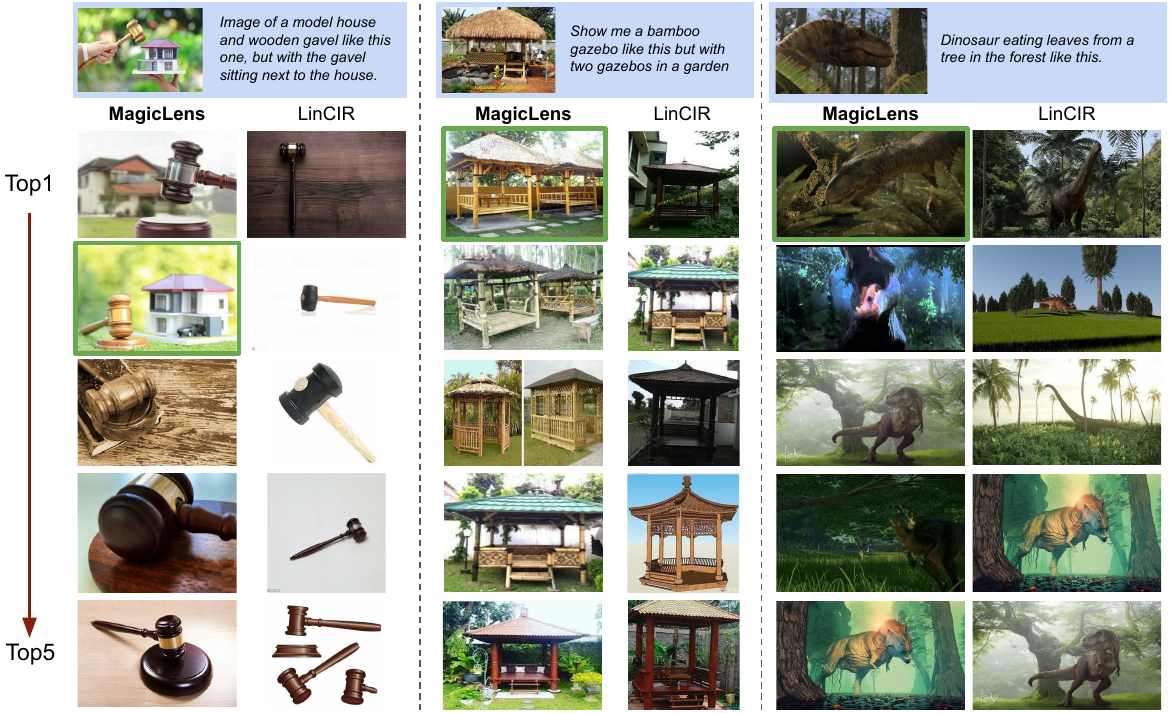}
    \caption{
    Top-5 retrieved images of CoCa-based \ModelName-L and LinCIR on the holdout index set with 1.4M images for queries shown in Figure~\ref{fig:case-study-holdout} and more.
    Queries are with a blue background and only the most correct retrieved images are marked with green outlines.
    }
    \label{appendix-fig:case-study-holdout-top5}
\end{figure}

We present detailed top-5 retrieval results of CoCa-based \ModelName-L and the code-available SOTA LinCIR~\cite{Gu2023LinCIR} in Figure~\ref{appendix-fig:case-study-holdout-top5}.
1) For the \texttt{bag} query, \ModelName can retrieve bags (the third and fourth images) from the same brand, even though they don't have shared visual clues (brand logo) with the query image.
2) Given the \texttt{house and gavel} query, our model successfully finds an interesting real-world scene and the perfect example in the top-2 results, but LinCIR fails to satisfy the query.
This may stem from the limited representation abilities of a single pseudo token for an image with multiple objects.
3) The success on the \texttt{gazebo} example shows that \ModelName can understand simple numerical relations.

\end{document}